%% file: main.tex
\documentclass[10pt,twocolumn,letterpaper]{article}

\usepackage{wacv}                         

\input{preamble}

\definecolor{wacvblue}{rgb}{0.21,0.49,0.74}
\usepackage[breaklinks,colorlinks,allcolors=wacvblue]{hyperref}

\def\wacvPaperID{98} 
\def\confName{WACV}
\def\confYear{2027}

\title{MAPA: Mode-Aware Prototype Allocation for Latent Dataset Distillation in 3D Medical Classification}

\author{Weiqi Zhang \qquad Hiroyuki Kubo\\
Chiba University, Japan\\
{\tt\small \{weiqizhang,hkubo\}@chiba-u.jp}}

\begin{document}
\maketitle
\input{ieee_notice}
\input{sec/0_abstract}
\input{sec/1_intro}
\input{sec/2_related_work}
\input{sec/3_method}
\input{sec/4_experiments}
\input{sec/5_conclusion}
\input{sec/6_acknowledgments}
{
    \small
    \bibliographystyle{ieeenat_fullname}

\input{main.bbl}
}

\clearpage
\setcounter{section}{0}
\setcounter{subsection}{0}
\setcounter{equation}{0}
\setcounter{figure}{0}
\setcounter{table}{0}
\setcounter{algorithm}{0}
\setcounter{footnote}{0}
\makeatletter
\def\theHsection{supp.\arabic{section}}
\def\theHsubsection{supp.\arabic{section}.\arabic{subsection}}
\def\theHequation{supp.\arabic{equation}}
\def\theHfigure{supp.\arabic{figure}}
\def\theHtable{supp.\arabic{table}}
\def\theHalgorithm{supp.\arabic{algorithm}}
\def\theHfootnote{supp.\arabic{footnote}}
\makeatother

\maketitlesupplementary
\raggedbottom
\makeatletter
\setlength{\@fptop}{0pt}
\setlength{\@dblfptop}{0pt}
\makeatother

\input{sec/supp_implementation}
\input{sec/supp_concentration}
\input{sec/supp_ae}
\input{sec/supp_significance}

\clearpage
\newlength{\supprefcolwidth}
\setlength{\supprefcolwidth}{\columnwidth}
\newlength{\supprefrightskip}
\setlength{\supprefrightskip}{\dimexpr\textwidth-\supprefcolwidth\relax}
\onecolumn
\begin{list}{}{\setlength{\leftmargin}{0pt}\setlength{\rightmargin}{\supprefrightskip}\setlength{\listparindent}{0pt}\setlength{\itemindent}{0pt}\setlength{\parsep}{0pt}}
\item[]
{
    \small
    \bibliographystyle{ieeenat_fullname}

\input{supplementary_refs.bbl}
}
\end{list}
\end{document}

%% file: preamble.tex
\usepackage{multirow}
\usepackage{array}
\usepackage{colortbl}
\usepackage{algorithm}
\usepackage{algpseudocode}

\makeatletter
\@ifpackageloaded{lineno}{%
  \setlength{\linenumbersep}{0.42cm}%
  \AtBeginEnvironment{figure}{\nolinenumbers}%
  \AtEndEnvironment{figure}{\linenumbers}%
  \AtBeginEnvironment{figure*}{\nolinenumbers}%
  \AtEndEnvironment{figure*}{\linenumbers}%
  \AtBeginEnvironment{table}{\nolinenumbers}%
  \AtEndEnvironment{table}{\linenumbers}%
  \AtBeginEnvironment{table*}{\nolinenumbers}%
  \AtEndEnvironment{table*}{\linenumbers}%
  \AtBeginEnvironment{algorithm}{\nolinenumbers}%
  \AtEndEnvironment{algorithm}{\linenumbers}%
}{}
\makeatother

\makeatletter
\renewcommand\paragraph{\@startsection{paragraph}{4}{\z@}%
  {0pt}%
  {-0.8em}%
  {\normalfont\normalsize\bfseries}}
\makeatother

\definecolor{oursrow}{RGB}{238,236,236}
\newcolumntype{C}[1]{>{\centering\arraybackslash}m{#1}}
\newcommand{\best}[1]{\textbf{#1}}
\newcommand{\second}[1]{\underline{#1}}
\newcommand{\maparow}{\rowcolor{oursrow}}
\newcommand{\ms}[2]{\mbox{\rmfamily #1\,\ensuremath{\pm}\,#2}}
\newcommand{\tnum}[1]{\mbox{\rmfamily #1}}

%% file: ieee_notice.tex
\AddToHookNext{shipout/foreground}{%
  \put(54,-776){%
    \parbox[b]{504pt}{%
      \normalfont\fontsize{7}{8}\selectfont
      \copyright\ 2027 IEEE. Personal use of this material is permitted.
      Permission from IEEE must be obtained for all other uses, in any current or future media,
      including reprinting/republishing this material for advertising or promotional purposes,
      creating new collective works, for resale or redistribution to servers or lists,
      or reuse of any copyrighted component of this work in other works.%
    }%
  }%
}

%% file: sec/0_abstract.tex
\begin{abstract}
Dataset distillation (DD) synthesizes a small training set from a large one,
but applying it to 3D medical classification is constrained by the memory and
optimization costs of synthesizing full volumetric samples. Under a
storage-equivalent budget, distilling in the latent space of a frozen
autoencoder reduces this cost and fits more synthetic samples (prototypes)
than in voxel space. However, class-level matching, such as distribution
matching, causes prototypes to concentrate in dense latent regions while
underrepresenting sparser intra-class regions, a consistent imbalance that we
term prototype concentration. To address this, we propose MAPA (Mode-Aware
Prototype Allocation), a latent distillation method that spreads prototypes
across intra-class modes while preserving class-level matching. MAPA marks
these modes with fixed anchors estimated by clustering the real latents and
distributes prototypes over them through a soft assignment, together with a lightweight
anti-collapse regularizer and curriculum weighting that stabilize the
optimization. Across five 3D medical classification tasks, MAPA improves the
aggregate balanced accuracy by +4.2 points over the strongest latent baseline
while using the same number of prototypes. An exploratory severity analysis
on MosMedData suggests improved worst-stratum recall without loss of normal-case specificity.
\end{abstract}

%% file: sec/1_intro.tex
\section{Introduction}
\label{sec:intro}

Dataset distillation (DD) aims to synthesize a small dataset from a large
one such that a model trained on the synthetic set approximates the
performance of a model trained on the full data~\cite{wang2018dd}. As a
data-efficient paradigm, DD has been studied
extensively~\cite{sachdeva2023survey, lei2024survey} and is especially
valuable for medical imaging, where datasets are costly to store, share,
and reuse across institutions and are constrained by privacy issues and
regulations~\cite{li2023medsharing}. Many medical imaging modalities, such as CT and MRI, produce 3D
volumetric data rather than 2D images, which makes distillation for 3D
medical data particularly relevant. Yet despite substantial progress on 2D
image classification~\cite{zhao2023dm, cazenavette2022mtt, liu2025evolution}
and growing interest in medical dataset distillation~\cite{li2025dataset},
DD for 3D medical classification remains largely limited to direct
voxel-space adaptations~\cite{yu2024ptm}.

\begin{figure}[t]
\centering
\includegraphics[width=\linewidth]{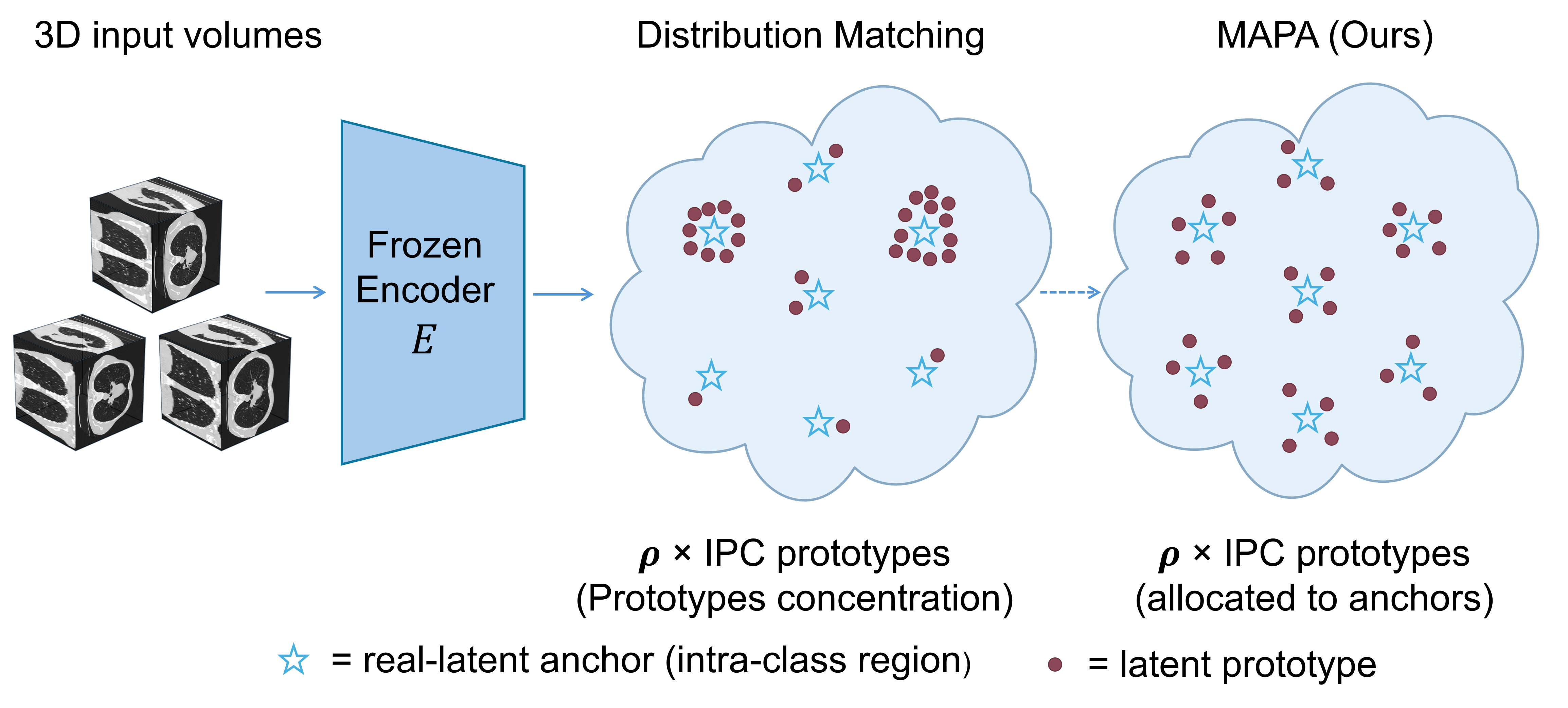}
\caption{Comparison of distribution matching~\cite{zhao2023dm}, a representative class-level
objective, and MAPA. With fixed real-latent anchors marking intra-class
modes, distribution matching can leave sparser modes underrepresented,
whereas MAPA spreads the same prototypes for broader coverage, which leads to
better learning performance for the downstream 3D medical imaging task.}
\label{fig:mapa_overview}
\end{figure}

At its core, DD optimizes synthetic data by matching a
learning signal between synthetic and real data, such as
gradients~\cite{zhao2021dc}, feature distributions~\cite{zhao2023dm}, or
training trajectories~\cite{cazenavette2022mtt}. This matching paradigm is
independent of where the synthetic data lives. The most direct way to extend
DD to 3D is to optimize synthetic volumes in voxel space. However, operating
on full 3D volumes is costly, as distillation must repeatedly process
high-dimensional tensors through memory-heavy 3D networks, incurring
prohibitive time and peak-memory cost at larger budgets or higher
resolutions~\cite{yu2024ptm}.

Latent dataset distillation moves the optimization out of voxel space,
optimizing compact latent prototypes in the low-dimensional space of a
frozen autoencoder~\cite{rombach2022ldm, duan2023latentdd} that are decoded
into volumes only for downstream training. Unlike a prototype in metric
learning, which denotes a single class mean~\cite{NIPS2017_cb8da676}, here a prototype can be one of several synthetic
latent vectors per class in the distilled set. Because a latent prototype is
far more compact than the volume it encodes, latent methods hold more
prototypes than their voxel-space counterparts~\cite{duan2023latentdd}
under the same storage budget, which we quantify at the byte level following
the storage-based comparison of dataset distillation~\cite{bao2025rateutility}.

However, while latent space DD can use additional prototypes, this does not
automatically result in improved learning performance.
Regardless of whether they match gradients, features, or trajectories, latent
methods relying on class-level matching tend to over-populate dense regions
while leaving other sparser
intra-class regions underrepresented. At low prototype counts, these sparser regions may receive
no prototype at all. We refer to this as
\textbf{prototype concentration}. This is
consequential in medical image classification, where sparser regions reflecting
anatomy, lesion extent, or hidden clinical
subgroups~\cite{DD46826, willemink2020preparing, oakdenrayner2020hidden}
may correspond to rare but important disease patterns that a diagnostic model
should not overlook.

To address prototype concentration that latent DD induces, we propose \textbf{M}ode-\textbf{A}ware
\textbf{P}rototype \textbf{A}llocation (MAPA).
For each class, MAPA estimates real-latent anchors, the K-means centroids of
the real training latents, which mark representative intra-class modes.
It keeps them active as coverage targets throughout distillation and
allocates prototypes across them (Figure~\ref{fig:mapa_overview}) while
preserving class-level matching. We
evaluate MAPA on five 3D medical classification tasks spanning diverse
modalities, anatomies, and
scales~\cite{DD46826, medmnistv1, medmnistv2, bien2018deep, MRNetdataset}.
In summary, our contributions are threefold. First, we present the first
systematic study of latent dataset distillation for 3D medical classification
and identify prototype concentration. Second, we propose MAPA, a latent
distillation method that mitigates prototype concentration through mode-aware
allocation over fixed real-latent anchors. Finally, we validate MAPA on five
3D medical classification tasks, where it attains the best
aggregate balanced accuracy among latent methods, improving the across-setting
aggregate balanced accuracy by $+4.2$ points over the strongest latent
baseline ($74.06$ versus\ $69.91$ mean bAcc) while using the same number of
prototypes. This aggregate evidence does not imply a significant improvement
in every setting. An exploratory severity analysis on MosMedData suggests
improved worst-stratum recall without loss of normal-case specificity (Table~\ref{tab:severity_recall}).

%% file: sec/2_related_work.tex
\section{Related Work}
\label{sec:related}

\subsection{Dataset Distillation}
\label{subsec:dd}

Dataset distillation (DD) synthesizes a small training set that approximates
the utility of a large dataset~\cite{wang2018dd}. Most classical methods
optimize synthetic samples directly in the input domain, typically in pixel
space. These methods differ in the learning signal matched between real and
synthetic data. Gradient matching
(DC)~\cite{zhao2021dc} aligns real and synthetic gradients, distribution
matching (DM)~\cite{zhao2023dm} matches feature-space statistics under
randomly sampled networks, and trajectory matching
(MTT)~\cite{cazenavette2022mtt} matches training trajectories, with
subsequent work improving these objectives through
augmentation~\cite{zhao2021dsa}, efficient
parameterization~\cite{kim2022idc}, attention
matching~\cite{sajedi2023datadam}, and scaling~\cite{cui2023tesla}. These
DD techniques are effective in 2D, but their cost grows rapidly with the sample
dimensionality, which makes direct voxel-space optimization difficult for 3D
volumes. In the medical domain, PTM~\cite{yu2024ptm} applies trajectory
matching to dataset distillation but still optimizes full volumes,
inheriting this cost.

To address this scalability concern, DD research has focused on developing
compact representations for the synthetic data.
Factorization-based methods decompose synthetic data into lower-dimensional
components~\cite{liu2022haba, lee2022kfs} but still restore them to the
high-dimensional pixel space at each optimization step and backpropagate through the
decoder. Latent dataset distillation~\cite{duan2023latentdd} instead moves
the distillation into the latent space of a frozen
autoencoder~\cite{rombach2022ldm}, optimizing prototypes, decoded
only for downstream training, which is attractive for 3D volumes where the
voxel-to-latent storage gap is large. While prior latent DD often
characterizes this gap through compression factors, we measure it in bytes
within our experimental protocol to make storage-equivalent comparisons
explicit~\cite{bao2025rateutility}. Our work extends latent DD methods by
determining how synthetic samples (prototypes) should be
allocated across intra-class regions given that the compact representation
allows for more prototypes than traditional DD techniques.

\subsection{Mode Coverage and Mode-Guided Distillation}
\label{subsec:mode}

Mode coverage is widely studied in generative modeling, where a generator may
capture only part of the real distribution, a failure known as mode
collapse~\cite{goodfellow2014gan}. This has
motivated diversity-promoting objectives such as mode-seeking
regularization~\cite{mao2019msgan} and clustering-based
coverage~\cite{liu2020selfcond}. Our notion of prototype concentration is
related in spirit but arises in a different setting, with a fixed number of
optimized prototypes rather than samples drawn from a generator. The question is not
whether a generator produces diverse samples, but how a fixed set of
prototypes is allocated across intra-class regions. This matters for 3D
medical imaging,
where volumetric data varies considerably within a class in anatomy, lesion
extent, and acquisition
condition~\cite{willemink2020preparing, oakdenrayner2020hidden, cpds2026review},
so that collapsing onto a few dominant modes leaves task-relevant variations
underrepresented.

Related coverage concerns appear in DD more broadly, motivating
objectives that promote diversity and discourage collapse onto dense
modes~\cite{gu2024minimax, sun2024rded, liu2025wasserstein}.
Diversity objectives encourage dispersion, whereas MAPA explicitly allocates
a fixed prototype set using estimates of modes within each class.
A separate line of research relies on pretrained
generative priors, such as GLaD~\cite{cazenavette2023glad} and the
mode-guided MGD$^3$~\cite{mgd3}, which synthesize samples through a large
pretrained model. MGD$^3$ uses intra-class modes to guide sample generation.
MAPA instead allocates a fixed prototype set across modes estimated by
real-latent anchors under a storage-equivalent budget, using the frozen autoencoder only as a fixed
encoder-decoder rather than a generative sampler. These methods require
training a generative model, so adapting them to 3D medical volumes would
demand substantial training time and resources. We therefore treat them as
conceptually related and primarily compare against optimization-based latent
distillation methods that do not require retraining a generative model on
medical imaging data.

%% file: sec/3_method.tex
\section{Method}
\label{sec:method}

\begin{figure*}[t]
\centering
\includegraphics[width=\textwidth]{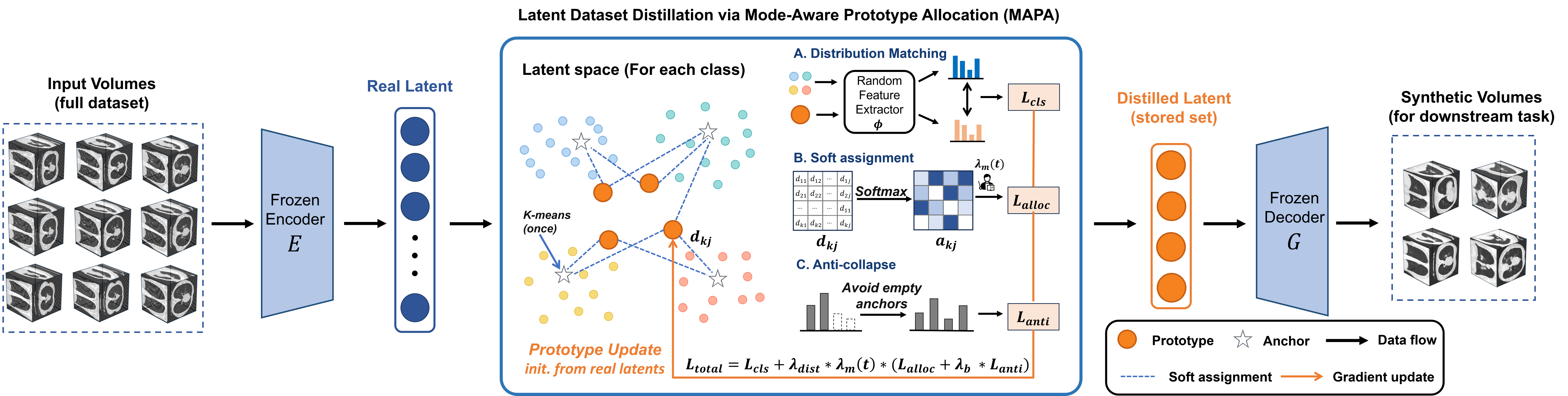}
\captionsetup{width=\textwidth}
\caption{MAPA optimization pipeline. A frozen encoder $E$ maps volumes to
latent space, where real-latent anchors (stars) mark estimated intra-class
modes. Latent prototypes (orange) are the stored distilled set, optimized by
class-level distribution matching ($\mathcal{L}_{\text{cls}}$), soft
prototype-to-anchor assignment ($\mathcal{L}_{\text{alloc}}$), and an
anti-collapse regularizer ($\mathcal{L}_{\text{anti}}$), then decoded by
frozen $G$ for downstream training. Blue denotes frozen or real components,
orange optimized prototypes.}
\label{fig:mapa_pipeline}
\end{figure*}

Figure~\ref{fig:mapa_pipeline} summarizes the MAPA optimization pipeline. It
combines class-level distribution matching ($\mathcal{L}_{\text{cls}}$), a
soft prototype-to-anchor assignment ($\mathcal{L}_{\text{alloc}}$), and an
anti-collapse regularizer ($\mathcal{L}_{\text{anti}}$), which we define
later in this section.

\subsection{Preliminaries}
\label{subsec:prelim}

\paragraph{Dataset distillation.}
Given a training set $\mathcal{D}=\{(x_i,y_i)\}_{i=1}^{|\mathcal{D}|}$ with
volumes $x_i\in\mathbb{R}^{D\times H\times W}$ and classification labels
$y_i\in\{1,\dots,C\}$, dataset distillation seeks a small synthetic set
$\mathcal{S}$ ($|\mathcal{S}|\ll|\mathcal{D}|$) such that a model trained on
$\mathcal{S}$ approximates a model trained on $\mathcal{D}$~\cite{wang2018dd}.
\begin{equation}
\mathcal{S}^\star = \arg\min_{\mathcal{S}} \mathcal{L}_{\mathcal{D}}(\theta^{\mathcal{S}})
\quad \text{subject to} \quad \theta^{\mathcal{S}}=\arg\min_\theta \mathcal{L}_{\mathcal{S}}(\theta).
\label{eq:dd}
\end{equation}
The budget is conventionally measured by images per class (IPC). Although our
data are 3D volumes rather than single images, we retain IPC as a metric for
comparability while enforcing storage equivalence at the byte
level~\cite{bao2025rateutility}.

\paragraph{Storage-equivalent latent prototype budget.}
A frozen autoencoder $(E,G)$ encodes a volume
$x\in\mathbb{R}^{D\times H\times W}$ into a latent representation
$z=E(x)\in\mathbb{R}^{c_z\times D'\times H'\times W'}$ with a spatial
downsampling factor $f$. Because the channel dimension grows from $1$ to
$c_z$, the byte-level storage ratio
\begin{equation}
\rho = \frac{\text{bytes}(x)}{\text{bytes}(z)}
\label{eq:rho}
\end{equation}
differs from $f$ and governs the storage-equivalent comparison. Under the
same byte budget as IPC voxel-space volumes per class, latent methods store
$\rho$ times as many prototypes, denoted $\{z_k^c\}_{k=1}^{N_c}$, where
$N_c$ is the number of prototypes per class. Following
byte-level storage accounting~\cite{bao2025rateutility}, we also account for
the decoder parameters, which are shared across all prototypes as a one-time
cost and are reported separately from the per-prototype storage
(Supplementary~Table~\ref{tab:supp_accounting}). The autoencoder is trained only on
the target training split and is then frozen, with its cost reported
separately, so our efficiency claims concern the distillation stage.

\paragraph{Class-level distribution matching.}
Our starting point is the class-level distribution-matching objective of
latent distillation~\cite{duan2023latentdd}. Under a 3D feature extractor
$\phi$, it aligns the mean of the
synthetic prototypes with the mean of the real latents of the same class.
\begin{equation}
\mathcal{L}_{\text{cls}}^c =
\Big\| \tfrac{1}{N_c}\sum_{k=1}^{N_c}\phi(z_k^c)
- \tfrac{1}{|\mathcal{D}_c|}\sum_{x_i\in\mathcal{D}_c}\phi(E(x_i)) \Big\|^2,
\label{eq:cls}
\end{equation}
where $\mathcal{D}_c=\{x_i:y_i=c\}$.
Following distribution matching~\cite{zhao2023dm}, $\phi$ is a randomly
initialized 3D network resampled at every distillation step and never trained.
This reduces dependence on a particular feature space and keeps the
protocol consistent across the compared distribution matching methods.
We do not study pretrained extractors in this work.
This matches the class mean but does
not constrain how the prototypes are distributed across intra-class regions.
In the low-budget regime, the additional prototypes can therefore be
absorbed by dense regions while sparser regions remain underrepresented. In
our comparisons, all latent baselines share the same prototype count, initialization,
number of distillation steps, and, where applicable, random-feature protocol as MAPA, so
performance differences reflect how the prototypes are allocated rather than
how many of them are used.

\subsection{Real-Latent Anchors}
\label{subsec:anchors}

A class typically spans several modes in the latent space, with dense
regions corresponding to distinct intra-class variations. For example, an
abnormal-knee class in an MRI dataset spans several abnormality types within
the same diagnostic label, rather than a single
appearance~\cite{bien2018deep, MRNetdataset}. To make this structure
explicit, we estimate a fixed set of real-latent anchors that mark these
modes, obtained by K-means on the real latents of the class.
\begin{equation}
\{\mu_j^c\}_{j=1}^{K} =
\operatorname*{arg\,min}_{\mu_1,\dots,\mu_K}
\sum_{x_i\in\mathcal{D}_c} \min_{j} \big\|E(x_i)-\mu_j\big\|_F^2,
\label{eq:anchors}
\end{equation}
with $\mu_j^c\in\mathbb{R}^{c_z\times D'\times H'\times W'}$ and $\|\cdot\|_F$
the Frobenius norm on the latent tensor. This is solved with Lloyd's
algorithm~\cite{lloyd1982least} once per class before distillation.
As cluster centroids, these anchors need not be the latent representation
of any individual training sample.
The anchors estimate these latent-space modes and are not assumed to
correspond to clinical subtypes. They are computed once before distillation
and kept frozen, as a one-time preprocessing step that is reused across all
IPC settings. Unlike a mere initialization, the anchors stay active
throughout distillation as coverage targets that the allocation objective
(Section~\ref{subsec:alloc}) optimizes against.

\paragraph{Choice of $K$.}
We set the anchor count $K$ per dataset from the intra-class heterogeneity
described in its documentation, and we fix it across all IPC settings. We use
a larger $K$ for datasets with greater expected intra-class appearance
variation and a smaller $K$ for more homogeneous tasks. MosMedData (chest
CT) shows the widest intra-class variation in disease extent and is
assigned $K{=}8$~\cite{DD46826}. Consistent with this, its positive cases
are clinically graded into several severity levels, although these grades are
not used as labels. The three MedMNIST3D subsets target more homogeneous
single-organ or single-lesion tasks and use $K{=}4$~\cite{medmnistv2}.
MRNet contrasts normal knees against a broad abnormal class and uses
$K{=}2$~\cite{bien2018deep, MRNetdataset}. These are coarse
estimates rather than tuned values, and we report sensitivity to
$K\in\{1,2,4,8,16\}$ in Section~\ref{subsec:sensitivity}. We further validate
these choices with a global $K$ sweep
(Supplementary~Table~\ref{tab:global_k_sweep}). $K$ is not tuned on the test set.
A fixed $K$ gives all methods the same anchors for the PCI comparison.
We use a fixed $K$ rather than density-based clustering such as
DBSCAN~\cite{ester1996density}, whose neighborhood radius is difficult to
set in high-dimensional latent spaces and whose cluster count is not controlled.
In the matched-IPC control, where
the raw
per-class count can fall below the default anchor count, we set the
effective anchor count to $K'=\min(K, N_c)$. The matched IPC${=}1$ case
therefore uses a single anchor per class, for which the allocation term
reduces to a single-anchor constraint and no longer tests multi-anchor
allocation.

\subsection{Allocating Prototypes across Intra-Class Modes}
\label{subsec:alloc}

Given the frozen anchors, which mark the estimated intra-class modes, MAPA
allocates its prototypes across them while preserving the class-level
matching of Equation~\eqref{eq:cls}. By distributing prototypes over these
anchors rather than only matching the class mean, the allocation becomes
mode-aware. It spreads capacity over the intra-class modes the anchors
estimate, instead of letting prototypes collapse onto the dominant regions.

\paragraph{Soft prototype-to-anchor assignment.}
Under the same feature extractor $\phi$ as in Equation~\eqref{eq:cls}, we measure
the feature-space distance between prototype $z_k^c$ and anchor $\mu_j^c$,
\begin{equation}
d_{kj}^c = \tfrac{1}{d_\phi}\big\|\phi(z_k^c)-\phi(\mu_j^c)\big\|^2,
\label{eq:dist}
\end{equation}
where $d_\phi$ is the feature dimension, and convert it into a soft
assignment through a temperature-controlled softmax,
\begin{equation}
a_{kj}^c =
\frac{\exp(-d_{kj}^c/\tau)}{\sum_{j'=1}^{K}\exp(-d_{kj'}^c/\tau)},
\quad \sum_{j=1}^{K} a_{kj}^c = 1,
\label{eq:assign}
\end{equation}
with the temperature $\tau{=}0.05$ fixed in all experiments. A soft assignment
keeps the mapping differentiable and lets each prototype be guided by
several nearby anchors. We compare against a hard nearest-anchor assignment
in Section~\ref{subsec:ablation}. The allocation loss pulls each prototype
toward its assigned anchors,
\begin{equation}
\mathcal{L}_{\text{alloc}}^c =
\tfrac{1}{N_c}\sum_{k=1}^{N_c}\sum_{j=1}^{K} a_{kj}^c\, d_{kj}^c.
\label{eq:alloc}
\end{equation}

\paragraph{Anti-collapse regularizer.}
Minimizing Equation~\eqref{eq:alloc} alone does not prevent the prototypes from
collapsing onto a few anchors. We therefore add a lightweight term that
penalizes concentrated aggregate assignment mass,
\begin{equation}
\mathcal{L}_{\text{anti}}^c =
\sum_{j=1}^{K}\Big(\tfrac{1}{N_c}\sum_{k=1}^{N_c} a_{kj}^c - \tfrac{1}{K}\Big)^2.
\label{eq:anti}
\end{equation}
Here $1/K$ is a collapse-free reference point rather than an assumption that
the true intra-class density is uniform. The term simply discourages
degenerate allocations in which most prototypes pile onto a single anchor.

\subsection{Overall Objective and Training}
\label{subsec:training}
\label{subsec:objective}

\paragraph{Objective.}
Combining class-level matching, allocation, and anti-collapse, the per-class
objective is
\begin{equation}
\mathcal{L}_{\text{total}}^c = \mathcal{L}_{\text{cls}}^c
+ \lambda_{\text{dist}}\,\lambda_m(t)\big(\mathcal{L}_{\text{alloc}}^c
+ \lambda_b\,\mathcal{L}_{\text{anti}}^c\big),
\label{eq:total}
\end{equation}
where $\lambda_{\text{dist}}$ scales the allocation terms relative to
class-level matching, $\lambda_b$ weights the anti-collapse term, and
$\lambda_m(t)$ is a curriculum weight over the distillation step $t$.

\paragraph{Curriculum weighting.}
Early in distillation, $\lambda_m(t)$ is small, so that class-level matching
dominates and the prototypes first align coarsely to the real distribution.
The allocation signal is then introduced gradually through a cosine schedule
up to $\lambda_m^{\max}$,
\begin{equation}
\lambda_m(t) =
\begin{cases}
\lambda_m^{\max}\cdot\tfrac{1}{2}\big(1-\cos(\pi t/0.7T)\big), & t \le 0.7T,\\[4pt]
\lambda_m^{\max}, & t > 0.7T,
\end{cases}
\label{eq:curriculum}
  \end{equation}
  where $T$ is the total number of distillation steps. This schedules the
allocation term over training time and is distinct from curricula over data
or augmentation~\cite{bengio2009curriculum, yin2023cda}. The temperature
$\tau$ is fixed throughout.

\paragraph{Optimization and hyperparameters.}
We optimize only the latent prototypes $\{z_k^c\}$ with Adam
(learning rate $5\times10^{-3}$, $T{=}100$ steps), keeping $(E,G)$ and every
sampled $\phi$ frozen. We fix $\lambda_{\text{dist}}{=}0.2$,
$\lambda_b{=}1$, and
$\tau{=}0.05$ across all experiments. The only per-dataset settings are the
anchor count $K$ (Section~\ref{subsec:anchors}) and the maximum allocation
weight $\lambda_m^{\max}$ ($0.03$ for MosMedData and MRNet, $0.10$ for
MedMNIST3D). The prototypes are initialized from randomly sampled real
latents of the corresponding class. The full hyperparameters and the
autoencoder configuration are given in
Supplementary~Tables~\ref{tab:supp_mapa_hparam_global}, \ref{tab:supp_mapa_hparam_data},
and~\ref{tab:supp_ae_config}. All hyperparameters,
including $K$ and $\lambda_m^{\max}$, are fixed a priori and none are selected on test
performance. The full training procedure is given in
Supplementary~Algorithm~\ref{alg:mapa}.

\paragraph{Prototype concentration diagnostic.}
To quantify concentration independently of any distillation method, we
assign each prototype to its nearest anchor,
$m_k^c=\arg\min_j\|z_k^c-\mu_j^c\|$, using fixed Euclidean distances in the
latent space rather than the resampled $\phi$, so that the scores are
comparable across methods. With the per-anchor occupancy
$\bar{a}_j^c=\tfrac{1}{N_c}\sum_k \mathbb{1}[m_k^c{=}j]$, where
$\mathbb{1}[\cdot]$ is the indicator function, and the entropy
$H(\bar{a}^c)=-\sum_j \bar{a}_j^c\log\bar{a}_j^c$, we define
\begin{equation}
\mathrm{PCI} = 1 - \frac{H(\bar{a})}{\log K}, \quad
\mathrm{EffModes} = \exp\big(H(\bar{a})\big),
\label{eq:pci}
\end{equation}
computed per class and then averaged over classes. PCI is $0$ when prototypes spread evenly and
approaches $1$ under heavy concentration, while EffModes lies in $[1,K]$.
Depending only on the distilled prototypes and the fixed anchors, PCI
applies to any latent distillation method and serves as a method-agnostic
diagnostic. It uses a hard nearest-anchor count for interpretability,
whereas the allocation objective uses the soft assignment of
Equation~\eqref{eq:assign}, so the diagnostic is not a restatement of the training
loss.

%% file: sec/4_experiments.tex
\begin{table*}[t]
\centering
\captionsetup{width=\textwidth}
\caption{Main MosMedData ($96^3$) comparison (bAcc/AUC mean$\pm$std over five seeds). The parenthesized size denotes dataset scale. Avg. is the mean over the three IPC settings and reports mean only. Dashes denote out-of-memory entries. Bold and underline denote the best and second-best results. The MAPA row is shaded. For bAcc, seed-paired MAPA gains over the strongest latent baselines have 95\% CIs of $[-1.24,+4.65]$, $[+0.89,+6.16]$, and $[-3.96,+4.54]$ at IPC $1$, $10$, and $20$, respectively.}
\label{tab:mosmed_main}
\scriptsize
\setlength{\tabcolsep}{4.8pt}
\renewcommand{\arraystretch}{1.08}
\resizebox{\textwidth}{!}{%
\begin{tabular}{@{}ll|cccc|cccc@{}}
\toprule
\multirow{2}{*}{Family} & \multirow{2}{*}{Method} &
\multicolumn{4}{c|}{bAcc} &
\multicolumn{4}{c}{AUC} \\
& & IPC=1 & IPC=10 & IPC=20 & Avg. & IPC=1 & IPC=10 & IPC=20 & Avg. \\
\midrule
Coreset & Random & \ms{53.75}{6.07} & \ms{58.14}{4.80} & \ms{60.98}{4.29} & 57.62 & \ms{57.47}{12.38} & \ms{60.43}{7.99} & \ms{67.08}{5.54} & 61.66 \\
 & KMeans & \ms{61.62}{0.77} & \ms{61.51}{3.04} & \ms{64.54}{5.38} & 62.56 & \ms{64.40}{0.12} & \ms{67.30}{2.56} & \ms{70.54}{8.30} & 67.41 \\
\midrule
\multirow{3}{*}{3D DD} & DC~\cite{zhao2021dc} & \ms{57.01}{4.86} & \ms{58.59}{4.09} & \ms{62.83}{2.78} & 59.48 & \ms{58.30}{10.94} & \ms{60.77}{6.70} & \ms{66.48}{7.07} & 61.85 \\
 & DM~\cite{zhao2023dm} & \ms{61.04}{6.28} & \ms{58.13}{5.05} & \ms{59.20}{7.49} & 59.46 & \ms{58.84}{11.48} & \ms{62.75}{5.68} & \ms{64.13}{8.76} & 61.91 \\
 & MTT~\cite{cazenavette2022mtt} & \ms{58.47}{5.67} & \ms{57.72}{7.60} & -- & -- & \ms{57.95}{12.27} & \ms{61.92}{8.82} & -- & -- \\
\midrule
\multirow{4}{*}{Latent DD} & MMD~\cite{JMLR:v13:gretton12a} & \ms{61.91}{3.70} & \ms{64.83}{6.99} & \ms{65.97}{7.39} & 64.24 & \second{\ms{66.36}{2.28}} & \ms{71.65}{7.59} & \ms{74.87}{4.06} & 70.96 \\
 & DC~\cite{zhao2021dc} & \ms{60.02}{3.12} & \ms{68.23}{3.50} & \ms{67.91}{2.25} & 65.39 & \ms{64.42}{2.42} & \ms{77.58}{1.69} & \ms{77.32}{3.13} & 73.11 \\
 & MTT~\cite{cazenavette2022mtt} & \second{\ms{62.69}{2.10}} & \ms{64.41}{4.60} & \second{\ms{71.78}{3.32}} & 66.29 & \ms{65.87}{1.31} & \ms{72.66}{4.30} & \second{\ms{79.29}{3.33}} & 72.61 \\
 & DM~\cite{zhao2023dm} & \ms{60.69}{1.74} & \second{\ms{69.64}{3.09}} & \ms{69.75}{5.13} & \second{66.69} & \ms{65.13}{1.93} & \second{\ms{79.63}{1.78}} & \ms{76.40}{6.04} & \second{73.72} \\
\midrule
\maparow Ours & \textbf{MAPA} & \best{\ms{64.92}{3.57}} & \best{\ms{73.42}{2.45}} & \best{\ms{73.03}{5.42}} & \best{70.46} & \best{\ms{71.11}{6.03}} & \best{\ms{81.69}{3.16}} & \best{\ms{79.35}{3.40}} & \best{77.38} \\
\midrule
\multicolumn{2}{c|}{Full-data ref.} & \multicolumn{8}{c@{}}{\ms{78.68}{4.01} / \ms{88.70}{2.30}} \\
\multicolumn{2}{c|}{Full-latent ref.} & \multicolumn{8}{c@{}}{\ms{69.55}{4.42} / \ms{78.16}{4.93}} \\
\bottomrule
\end{tabular}
}
\end{table*}

\section{Experiments}
\label{sec:exp}
\vspace{-0.35\baselineskip}

\subsection{Setup and Implementation Details}
\label{subsec:setup}

\paragraph{Datasets.}
We evaluate on five 3D medical classification tasks spanning different
modalities, anatomies, and scales, namely MosMedData~\cite{DD46826} (chest CT,
COVID-19 versus\ normal, $96^3$), three MedMNIST3D subsets~\cite{medmnistv1, medmnistv2}
(OrganMNIST3D, NoduleMNIST3D, VesselMNIST3D, $28^3$), and
MRNet~\cite{bien2018deep, MRNetdataset} (knee MRI, abnormality
detection, $128^3$). These tasks cover binary and multi-class settings.

\paragraph{Autoencoder and storage-equivalent budget.}
For each dataset we train a lightweight 3D autoencoder on its training
split and freeze it for distillation (Section~\ref{subsec:prelim}).
Using the byte-level protocol of Section~\ref{subsec:prelim}, the
storage-equivalent prototype count is $\mathrm{IPC}\cdot\rho$. In our
datasets, $28^3$, $96^3$, and $128^3$ volumes correspond to $\rho{=}4$,
$32$, and $64$, respectively, with geometry specified in
Supplementary~Table~\ref{tab:supp_budget}.

\paragraph{Baselines.}
Coreset baselines use random and K-means selection of real volumes. Voxel-space DD
uses our 3D adaptations of DC~\cite{zhao2021dc}, DM~\cite{zhao2023dm}, and
MTT~\cite{cazenavette2022mtt}. The MTT variant uses progressive trajectory
matching rather than a naive 3D port, with PTM~\cite{yu2024ptm} as the
closest benchmark reference. Latent DD includes latent-space DC, DM, MTT,
and MMD~\cite{JMLR:v13:gretton12a} (maximum mean discrepancy) under the same storage-equivalent budget as MAPA. We also report full-data and
full-latent references as reference points for the voxel and latent regimes.

\paragraph{Protocol.}
Distilled sets are evaluated by training a 3D ConvNet
classifier~\cite{tran2015learning} from scratch and reporting balanced
accuracy (bAcc) and AUC. For MosMedData we use IPC $\in\{1,10,20\}$ and for
the extension datasets IPC $\in\{1,10\}$. Unless otherwise noted, all
results are averaged over five random seeds (42 to 46) and reported as mean $\pm$
standard deviation, and all latent methods share the same autoencoder,
prototype count, initialization, and protocol (Section~\ref{subsec:prelim}),
differing only in the distillation objective. Further implementation details are in
Supplementary~Section~\ref{sec:supp_impl}. All distillation-stage timings and peak-memory measurements
use a single NVIDIA RTX PRO 6000 Blackwell GPU (96\,GB GDDR7).
Code is available at \url{https://github.com/WeiqiZhang1999/MAPA}.

\paragraph{Fairness and budget protocol.}
Our primary storage-equivalent setting fixes MAPA and all latent baselines
at $\mathrm{IPC}\cdot\rho$ prototypes to isolate allocation. The matched IPC
control (Section~\ref{subsec:matchedipc}) verifies that MAPA gains do not stem
from more samples.

\subsection{Main Results}
\label{subsec:main}

Table~\ref{tab:mosmed_main} reports MosMedData results. MAPA attains the
best mean bAcc and AUC at every IPC, improving over the strongest latent
baseline by $+2.23$, $+3.78$, and $+1.25$ mean bAcc at IPC $1$, $10$, and
$20$, respectively. Gains are largest and most consistent at IPC${=}10$,
while IPC${=}1$ and IPC${=}20$ show positive mean margins with larger
seed-to-seed variation. At IPC${=}10$, MAPA reaches $73.42$ bAcc and $81.69$ AUC, above
the full-latent reference ($69.55/78.16$). This reference trains on all real latents without distillation and is not a
theoretical upper bound. A distilled set may improve generalization through regularization and
balanced class representation. MAPA reaches its strongest performance at IPC${=}10$ and plateaus at
IPC${=}20$. Its IPC${=}10$ accuracy exceeds all latent baselines at
IPC${=}20$, indicating that effective allocation drives the gain despite
using fewer prototypes. On MosMedData, latent methods outperform voxel distillation, which trails the
K-means coreset, and MAPA performs best. Under the byte-equivalent budget,
voxel methods lack the $\rho$-fold latent expansion and optimize fewer samples
directly in high dimensional 3D space, destabilizing matching and allowing real
coresets to outperform poorly optimized synthetic volumes. 3D-MTT runs out of
memory at IPC${=}20$.

Table~\ref{tab:extension_main} extends the comparison to four datasets.
MAPA has the best mean balanced accuracy in all settings, including MRNet,
where 3D-DC and 3D-MTT run out of memory at IPC${=}10$ on the 96\,GB GPU and
the feasible 3D-DM trails MAPA by a wide margin ($62.91$ versus\ $80.63$). On
OrganMNIST3D, MAPA approaches the full-data reference ($93.37$ versus\ $93.42$),
and on MRNet it exceeds both references ($80.63$ versus\ full-data
$77.54$ and full-latent $76.55$). These references are not upper bounds under our protocol, and the MRNet
full-data result varies substantially across seeds. Overall, the allocation benefit generalizes
across modalities, anatomies, and scales. Across all 11 storage-equivalent
latent settings, MAPA has the best mean bAcc and average rank, ranking first
in all 11 (Supplementary~Table~\ref{tab:supp_aggregate_rank}). Thus the improvement
is best characterized as a consistent aggregate accuracy and rank gain, not
a uniformly significant margin at every setting. Seed-paired bootstrap 95\%
intervals exclude zero for MosMedData IPC${=}10$ and OrganMNIST3D IPC${=}10$,
while the remaining positive mean margins, including NoduleMNIST3D
IPC${=}10$, have intervals that include zero.

\begin{table*}[!t]
\centering
\captionsetup{width=\textwidth}
\caption{Extension results (balanced accuracy). Parenthesized sizes denote dataset scale. Dashes denote out-of-memory entries. Bold and underline denote the best and second-best results. The MAPA row is shaded.}
\label{tab:extension_main}
\scriptsize
\setlength{\tabcolsep}{3.2pt}
\resizebox{\linewidth}{!}{%
\begin{tabular}{@{}ll|cc|cc|cc|cc@{}}
\toprule
\multirow{2}{*}{Family} & \multirow{2}{*}{Method} &
\multicolumn{2}{c|}{OrganMNIST3D ($28^3$)} & \multicolumn{2}{c|}{NoduleMNIST3D ($28^3$)} & \multicolumn{2}{c|}{VesselMNIST3D ($28^3$)} & \multicolumn{2}{c}{MRNet ($128^3$)} \\
\cmidrule(lr){3-4}\cmidrule(lr){5-6}\cmidrule(lr){7-8}\cmidrule(lr){9-10}
& & IPC$=1$ & IPC$=10$ & IPC$=1$ & IPC$=10$ & IPC$=1$ & IPC$=10$ & IPC$=1$ & IPC$=10$ \\
\midrule
Coreset & Random & \ms{50.72}{2.31} & \ms{87.62}{2.10} & \ms{50.63}{2.89} & \ms{62.30}{3.10} & \ms{60.44}{6.98} & \ms{67.59}{6.63} & \ms{55.20}{8.05} & \ms{61.49}{8.84} \\
 & KMeans & \ms{48.09}{0.98} & \ms{89.63}{1.17} & \ms{50.00}{0.00} & \ms{68.49}{2.99} & \second{\ms{63.50}{0.00}} & \ms{67.32}{2.23} & \ms{46.95}{0.00} & \ms{57.52}{5.35} \\
\midrule
\multirow{3}{*}{3D DD} & DC~\cite{zhao2021dc} & \ms{50.78}{3.43} & \ms{86.74}{2.47} & \ms{54.53}{5.48} & \ms{62.06}{6.42} & \ms{56.84}{6.38} & \ms{68.30}{4.16} & \ms{53.91}{5.59} & -- \\
 & DM~\cite{zhao2023dm} & \ms{50.68}{3.11} & \ms{88.09}{1.02} & \second{\ms{54.71}{7.32}} & \ms{60.50}{8.45} & \ms{58.26}{7.39} & \ms{69.70}{1.79} & \ms{52.86}{6.01} & \ms{62.91}{5.39} \\
 & MTT~\cite{cazenavette2022mtt} & \ms{49.42}{1.62} & \ms{86.83}{1.39} & \ms{54.46}{6.84} & \ms{60.04}{6.23} & \ms{59.30}{6.08} & \ms{68.87}{3.50} & \ms{53.79}{5.79} & -- \\
\midrule
\multirow{4}{*}{Latent DD} & MMD~\cite{JMLR:v13:gretton12a} & \ms{78.04}{3.41} & \ms{90.56}{2.60} & \ms{54.66}{5.38} & \second{\ms{71.56}{4.61}} & \ms{59.35}{6.19} & \second{\ms{74.13}{5.03}} & \ms{69.87}{4.32} & \ms{77.94}{3.20} \\
 & DC~\cite{zhao2021dc} & \ms{76.39}{3.68} & \ms{90.94}{1.67} & \ms{51.58}{4.44} & \ms{71.35}{4.29} & \ms{56.86}{3.93} & \ms{73.28}{2.89} & \ms{64.06}{2.08} & \ms{77.37}{1.88} \\
 & MTT~\cite{cazenavette2022mtt} & \ms{76.44}{5.08} & \ms{91.29}{1.75} & \ms{50.53}{3.99} & \ms{68.92}{2.90} & \ms{58.53}{5.47} & \ms{73.74}{2.81} & \ms{62.02}{5.43} & \ms{76.61}{3.01} \\
 & DM~\cite{zhao2023dm} & \second{\ms{78.40}{2.63}} & \second{\ms{91.48}{1.71}} & \ms{52.32}{3.78} & \ms{70.94}{5.79} & \ms{57.49}{4.18} & \ms{71.11}{5.78} & \second{\ms{69.20}{6.66}} & \second{\ms{77.94}{2.91}} \\
\midrule
\maparow Ours & \textbf{MAPA} & \best{\ms{82.36}{1.05}} & \best{\ms{93.37}{0.95}} & \best{\ms{62.53}{1.50}} & \best{\ms{72.05}{2.62}} & \best{\ms{66.07}{1.67}} & \best{\ms{75.57}{3.99}} & \best{\ms{70.76}{5.12}} & \best{\ms{80.63}{1.35}} \\
\midrule
\multicolumn{2}{c|}{Full-data ref.} & \multicolumn{2}{c|}{\ms{93.42}{1.01}} & \multicolumn{2}{c|}{\ms{80.73}{1.98}} & \multicolumn{2}{c|}{\ms{83.12}{1.17}} & \multicolumn{2}{c}{\ms{77.54}{5.05}} \\
\multicolumn{2}{c|}{Full-latent ref.} & \multicolumn{2}{c|}{\ms{93.07}{1.69}} & \multicolumn{2}{c|}{\ms{78.13}{2.68}} & \multicolumn{2}{c|}{\ms{84.44}{2.91}} & \multicolumn{2}{c}{\ms{76.55}{3.28}} \\
\bottomrule
\end{tabular}
}
\end{table*}

\paragraph{Latent selection versus optimization.}
A selection-only control shows that Latent-Medoids does not match MAPA, so
the gain requires optimized allocation rather than selecting real latents
(Supplementary~Section~\ref{subsec:supp_latent_selection}).

\paragraph{Severity-stratified performance.}
MosMedData provides CT severity grades for COVID-positive cases, allowing
an exploratory stratified analysis. Table~\ref{tab:severity_recall} compares representative
coreset, voxel-space, latent, and MAPA results at IPC${=}10$ from separate
ConvNet3D evaluations. These are not a stratification of the predictions
used for Table~\ref{tab:mosmed_main}. MAPA gives the highest COVID bAcc, COVID+ recall,
worst-stratum recall, severity-macro recall, and CT-0 specificity, improving
over the strongest non-MAPA baseline by $+13.6$, $+14.3$, $+15.9$, $+23.1$,
and $+6.3$ points, respectively.
These results suggest higher positive recall across severity strata
without loss of normal-case specificity. The small strata and high
variation across seeds make this exploratory evidence, rather than a
precise or uniformly significant effect.

\begin{table}[!t]
\centering
\caption{Exploratory severity evaluation on MosMedData at IPC=10, conducted
separately from Table 1. COVID bAcc averages the overall COVID+ recall and CT-0
specificity. Worst-stratum and severity-macro recall are the minimum and
unweighted mean across CT-1 to CT-4. Test counts for CT-0 to CT-4 are 51,
137, 23, 10, and 1. The single CT-4 case warrants caution. Bold indicates
the best result, and the MAPA row is shaded.}
\label{tab:severity_recall}
\scriptsize
\setlength{\tabcolsep}{3.0pt}
\resizebox{\linewidth}{!}{%
\begin{tabular}{lccccc}
\toprule
Method & COVID bAcc & COVID+ rec. & Worst rec. & Sev.-macro rec. & CT-0 spec. \\
\midrule
Volume-Random & \ms{58.39}{6.14} & \ms{48.54}{25.02} & \ms{41.47}{29.50} & \ms{58.91}{30.73} & \ms{68.24}{14.44} \\
3D-DM~\cite{zhao2023dm} & \ms{58.59}{6.06} & \ms{46.20}{20.36} & \ms{31.90}{30.06} & \ms{52.21}{30.83} & \ms{70.98}{8.81} \\
Latent-DM~\cite{zhao2023dm} & \ms{56.83}{6.09} & \ms{38.36}{35.25} & \ms{36.93}{34.53} & \ms{45.18}{41.44} & \ms{75.29}{24.55} \\
\maparow \textbf{MAPA} & \best{\ms{72.19}{3.46}} & \best{\ms{62.81}{7.95}} & \best{\ms{57.37}{8.86}} & \best{\ms{82.00}{5.27}} & \best{\ms{81.57}{11.65}} \\
\bottomrule
\end{tabular}
}
\end{table}

\subsection{Prototype Concentration}
\label{subsec:concentration}

Table~\ref{tab:mosmed_pci} quantifies prototype concentration across latent
methods under the same anchors and protocol. At IPC${=}10$, all class-level
matching baselines show similar concentration (PCI $\approx 4\%$), whether
they match gradients (DC), features (DM, MMD), or trajectories (MTT).
Prototype concentration is therefore a consequence of class-level matching
rather than any particular matching criterion. MAPA reduces PCI relative to
the Latent-DM baseline ($4.30\%\rightarrow1.70\%$) and raises normalized occupied modes to $96.62\%$
($7.73$ of $8$). Per-anchor prototype allocation for Latent-DM and MAPA is
shown in Supplementary~Figure~\ref{fig:mode_coverage}. PCI decreases and normalized
EffModes increase on all five datasets, although the absolute concentration
level varies with task difficulty, for example OrganMNIST3D retains higher PCI
(Supplementary~Table~\ref{tab:pci_all_datasets}). Coverage improves across datasets despite differences in absolute
concentration.

\begin{table}[!t]
\centering
\captionsetup{skip=2pt}
\caption{Prototype concentration on MosMedData latent methods. PCI and
normalized Eff. (\%) are defined only for latent methods.
Bold denotes the best result, and the MAPA row is shaded.}
\label{tab:mosmed_pci}
\scriptsize
\setlength{\tabcolsep}{2.4pt}
\renewcommand{\arraystretch}{0.88}
\resizebox{\linewidth}{!}{%
\begin{tabular}{l|cc|cc|cc}
\toprule
\multirow{2}{*}{Method} &
\multicolumn{2}{c|}{IPC=1} &
\multicolumn{2}{c|}{IPC=10} &
\multicolumn{2}{c}{IPC=20} \\
& PCI$\downarrow$ & Eff.$\uparrow$ & PCI$\downarrow$ & Eff.$\uparrow$ & PCI$\downarrow$ & Eff.$\uparrow$ \\
\midrule
MMD~\cite{JMLR:v13:gretton12a} & \ms{6.75}{3.98} & \ms{87.17}{7.05} & \ms{3.87}{1.90} & \ms{92.33}{3.61} & \ms{3.52}{1.66} & \ms{92.99}{3.20} \\
DC~\cite{zhao2021dc} & \ms{7.03}{3.79} & \ms{86.63}{6.69} & \ms{3.94}{1.83} & \ms{92.20}{3.47} & \ms{3.57}{1.62} & \ms{92.89}{3.12} \\
MTT~\cite{cazenavette2022mtt} & \ms{7.15}{4.24} & \ms{86.48}{7.38} & \ms{3.88}{1.90} & \ms{92.38}{3.62} & \ms{3.51}{1.67} & \ms{93.00}{3.21} \\
DM~\cite{zhao2023dm} & \ms{7.15}{4.24} & \ms{86.48}{7.38} & \ms{4.30}{2.90} & \ms{91.62}{5.38} & \ms{3.51}{1.67} & \ms{93.00}{3.21} \\
\maparow \textbf{MAPA} & \best{\ms{2.43}{3.28}} & \best{\ms{95.27}{6.08}} & \best{\ms{1.70}{1.70}} & \best{\ms{96.62}{3.38}} & \best{\ms{1.92}{2.10}} & \best{\ms{96.17}{4.17}} \\
\bottomrule
\end{tabular}
}
\end{table}

Baselines cover most anchors, with EffModes around $92\%$ or $7.3$ of $8$,
indicating moderate imbalance in which dominant anchors absorb disproportionate
mass while the budget remains usable. MAPA raises EffModes
($92\%\rightarrow96.62\%$) and lowers PCI relative to Latent-DM
($4.30\%\rightarrow1.70\%$). PCI is a distribution diagnostic rather than a
direct accuracy predictor, since accuracy also depends on the matching objective
and differs among baselines with similar PCI. Component and anchor source
ablations attribute the gain to explicit allocation rather than PCI reduction itself.

\subsection{Component Ablation}
\label{subsec:ablation}

Table~\ref{tab:ablation} ablates MAPA on MosMedData (IPC${=}10$), starting
from the prototype concentration baseline (Latent-DM). Hard nearest-anchor
allocation slightly hurts performance ($-0.88$), since committing each
prototype to a single anchor removes soft-assignment guidance. The
differentiable soft allocation improves over the baseline ($+2.15$), and
full MAPA, adding anti-collapse stabilization and curriculum weighting,
reaches the best result ($+3.78$). Removing the curriculum weighting alone
reduces the gain to $+0.90$, showing that gradually introducing the
allocation signal stabilizes optimization. Under a fixed matching objective, allocation reduces concentration and
improves accuracy over the same baseline.

\begin{table}[!t]
\centering
\captionsetup{skip=2pt}
\caption{Unified ablation on MosMedData (IPC=10). Columns Anchor, Alloc.,
Stabil., and Curric. indicate the anchor source, the assignment type, the
anti-collapse regularizer, and the curriculum weighting, with \checkmark
for enabled and -- for disabled. Gain ($\Delta$bAcc) is over
Latent-DM~\cite{zhao2023dm}. Bold denotes the best result, and the MAPA row
is shaded.}
\label{tab:ablation}
\begingroup
\scriptsize
\setlength{\tabcolsep}{2.4pt}
\renewcommand{\arraystretch}{0.88}
\resizebox{\linewidth}{!}{%
\begin{tabular}{@{}lcccccc@{}}
\toprule
Variant & Anchor & Alloc. & Stabil. & Curric. & bAcc & $\Delta$bAcc \\
\midrule
Prototype concentration baseline & -- & -- & -- & -- & \ms{69.64}{3.09} & \tnum{0.00} \\
\midrule
\multicolumn{7}{@{}l}{\textit{Allocation components}} \\
Hard allocation & Real & Hard & \checkmark & \checkmark & \ms{68.76}{3.29} & \tnum{-0.88} \\
Differentiable allocation & Real & Soft & -- & \checkmark & \ms{71.79}{3.93} & \tnum{+2.15} \\
MAPA without curriculum weighting & Real & Soft & \checkmark & -- & \ms{70.54}{4.15} & \tnum{+0.90} \\
\midrule
\multicolumn{7}{@{}l}{\textit{Anchor source}} \\
Random anchors & Random & Soft & \checkmark & \checkmark & \ms{69.86}{1.79} & \tnum{+0.22} \\
\maparow \textbf{MAPA} & Real & Soft & \checkmark & \checkmark & \best{\ms{73.42}{2.45}} & \best{\tnum{+3.78}} \\
\bottomrule
\end{tabular}
}
\endgroup
\end{table}

\paragraph{Anchor source.}
Table~\ref{tab:ablation} tests whether the gain depends on real intra-class
anchoring by replacing K-means real-latent anchors with random anchors,
otherwise keeping MAPA unchanged. Random anchors preserve the reference set
structure but remove real intra-class semantics, remaining near the Latent-DM
baseline ($69.86$ versus\ $69.64$ bAcc), while MAPA with real-latent anchors achieves
$73.42$. This attributes the gain to allocation toward real intra-class regions
rather than an arbitrary reference set.

\subsection{Sensitivity and Scaling}
\label{subsec:sensitivity}

\paragraph{Number of anchors.}
Supplementary~Figure~\ref{fig:mapa_k_ablation} reports sensitivity to
$K\in\{1,2,4,8,16\}$ on MosMedData. Performance varies moderately rather than
monotonically, with $K{=}8$ a strong default on MosMedData. Across all 11
settings, the dataset defaults match or exceed Latent-DM. When $K$ departs strongly from the dataset intra-class
variation, the performance gain weakens on the dataset, consistent with $K$ reflecting that variation
rather than acting as a tuned hyperparameter
(Supplementary~Table~\ref{tab:global_k_sweep}).

\paragraph{Cross-architecture readout.}
MAPA also remains the strongest on two alternative downstream architectures,
ResNet18-3D~\cite{tran2018closer} and
MobileNetV2-3D~\cite{Sandler_2018_CVPR}, with full results in
Supplementary~Section~\ref{subsec:supp_cross_arch}.

\paragraph{Prototype-count scaling.}
MAPA benefits more than the latent baselines from a larger prototype count,
confirming that extra prototypes help only when allocated effectively
(Supplementary~Section~\ref{sec:supp_proto_count_scaling}).

\subsection{Efficiency and Feasibility}
\label{subsec:efficiency}

\begin{figure}[t]
\centering
\includegraphics[width=\linewidth]{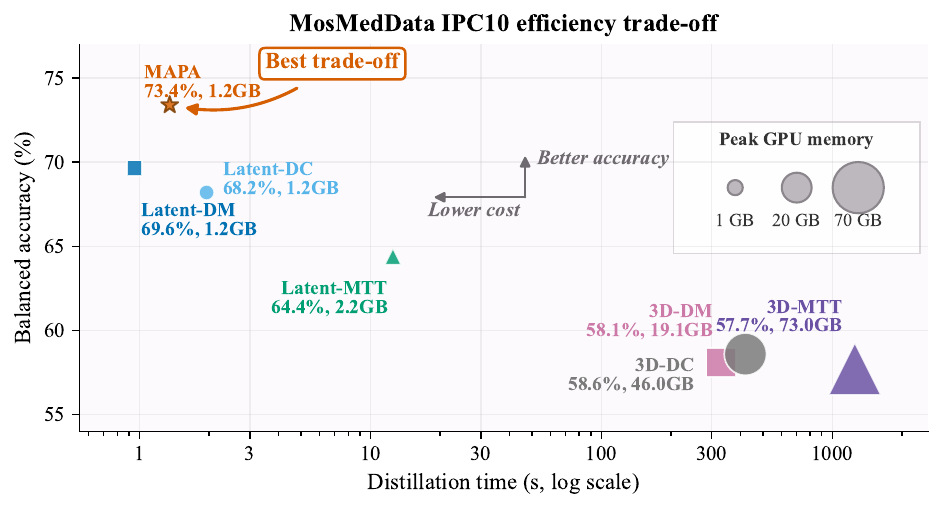}
\caption{Efficiency trade-off on MosMedData (IPC${=}10$). The axes show
distillation time in seconds on a log scale and balanced accuracy in
percent, and bubble size denotes peak GPU memory. Latent methods, including
MAPA, distill with a small peak-memory footprint, one to two orders of
magnitude below voxel-space methods, which makes single-GPU 3D distillation
feasible. Within this small memory budget, MAPA reaches the highest balanced
accuracy, adding only a lightweight allocation step over latent baselines.}
\label{fig:efficiency_tradeoff}
\end{figure}

Single GPU feasibility comes from the latent
paradigm~\cite{duan2023latentdd}. Figure~\ref{fig:efficiency_tradeoff} compares
distillation time and peak memory on MosMedData (IPC${=}10$). During
distillation, latent methods use one to two orders of magnitude less peak GPU
memory than voxel methods, enabling distillation beyond voxel memory limits.
The latter exhaust the 96\,GB GPU for 3D-MTT at IPC${=}20$ and for 3D-DC
and 3D-MTT on MRNet at IPC${=}10$.

This feasibility reduces peak memory and optimization difficulty but requires
longer downstream training on $\rho$ times as many decoded volumes under the
storage-equivalent budget. Including
the one-time autoencoder training cost, total time can exceed voxel distillation
(Supplementary~Table~\ref{tab:supp_accounting}). MAPA adds only a lightweight allocation
step ($1.32$\,s versus\ $1.01$\,s for Latent-DM at IPC${=}10$), leaving the
latent distillation efficiency profile largely unchanged.

\subsection{Matched-IPC Comparison}
\label{subsec:matchedipc}

As a stringent control, we force latent methods to use the same raw
per-class count as voxel methods (matched IPC of 1 and 10) rather than
their storage-equivalent budget.
This control tests whether MAPA gains depend on optimizing more samples
(Supplementary~Table~\ref{tab:mosmed_matched_ipc}). At matched IPC 1, MAPA shows no
advantage because the control necessarily uses $K'{=}1$ anchor per class
(Section~\ref{subsec:anchors}), leaving the multi-anchor mechanism inactive. At
matched IPC 10, multiple anchors are available and MAPA becomes competitive
again, confirming that its main-setting gains stem from allocation rather than
from more samples.

%% file: sec/5_conclusion.tex
\section{Conclusion}
\label{sec:conclusion}

We studied dataset distillation for 3D medical classification in the latent
space of a frozen autoencoder, where a storage-equivalent budget admits more
prototypes than voxel space. We found that class-level matching leaves these
prototypes concentrated in dense latent regions, an imbalance we term
prototype concentration and quantify with a method-agnostic index (PCI). To
address it, MAPA marks intra-class modes with fixed real-latent anchors and
allocates prototypes over them while preserving class-level matching. Across
five 3D medical classification tasks,
MAPA achieves the best aggregate balanced accuracy among latent methods under
the same prototype count while reducing concentration. An exploratory
analysis further suggests improved worst-stratum recall across the COVID-19
severity grades without loss of normal-case specificity. The method
relies on a frozen autoencoder trained separately for each dataset, so its
results are conditional on that latent representation. Each new dataset
therefore requires training another autoencoder. Other limitations are
clustering-based mode estimates without clinical subtype validation and
an anchor count set a priori from dataset documentation. Future work includes extending MAPA
to regression tasks.

%% file: sec/6_acknowledgments.tex
\paragraph{Acknowledgments.}
We thank Suren Jayasuriya for his guidance and feedback.
This work was supported by JSPS KAKENHI JP24K02953 and JST FOREST
JPMJFR206I.

%% file: sec/supp_implementation.tex
\section{Implementation Details}
\label{sec:supp_impl}

Algorithm~\ref{alg:mapa} summarizes MAPA in the standard distillation
form. The anchors are computed once from real latents and then frozen.
During distillation, losses are accumulated over classes and only the
latent prototypes are updated.

\begin{algorithm}[H]
\small
\caption{Mode-Aware Prototype Allocation (MAPA)}
\label{alg:mapa}
\begin{algorithmic}[1]
\Statex \textbf{Input} real dataset $\mathcal{D}=\bigcup_c\mathcal{D}_c$, frozen autoencoder $(E,G)$,
\Statex \hspace{\algorithmicindent} prototypes/class $N_c$, anchors/class $K$, iterations $T$, learning rate $\eta$,
\Statex \hspace{\algorithmicindent} weights $\lambda_{\mathrm{dist}},\lambda_b,\lambda_m^{\max}$, temperature $\tau$
\Statex \textbf{Output} distilled latent prototypes $\mathcal{S}=\{z_k^c\}$
\Statex
\Statex \textit{Preprocessing (once, frozen thereafter)}
\For{each class $c$}
  \State $r_i \gets E(x_i)$ for all $x_i\in\mathcal{D}_c$
  \State $\{\mu_j^c\}_{j=1}^{K}\gets\textsc{KMeans}(\{r_i\},K)$ \Comment{anchors, Equation~(4)}
  \State Initialize $\{z_k^c\}_{k=1}^{N_c}$ from real latents of class $c$
\EndFor
\Statex
\Statex \textit{Distillation}
\For{$t=1$ to $T$}
  \State Sample a random feature extractor $\phi$
  \State $\lambda_m \gets \lambda_m(t)$ \Comment{curriculum, Equation~(10)}
  \State $\mathcal{L}\gets0$
  \For{each class $c$}
    \State Compute class-level matching $\mathcal{L}_{\mathrm{cls}}^c$ \Comment{Equation~(3)}
    \State Compute prototype-anchor distances $d_{kj}^c$ \Comment{Equation~(5)}
    \State $a_{kj}^c\gets\mathrm{softmax}_j(-d_{kj}^c/\tau)$ \Comment{Equation~(6)}
    \State Compute $\mathcal{L}_{\mathrm{alloc}}^c$ and $\mathcal{L}_{\mathrm{anti}}^c$ \Comment{Equations~(7),(8)}
    \State $\mathcal{L}^c\gets\mathcal{L}_{\mathrm{cls}}^c+\lambda_{\mathrm{dist}}\lambda_m(\mathcal{L}_{\mathrm{alloc}}^c+\lambda_b\mathcal{L}_{\mathrm{anti}}^c)$ \Comment{Equation~(9)}
    \State $\mathcal{L}\gets\mathcal{L}+\mathcal{L}^c$
  \EndFor
  \State Update $\mathcal{S}\gets\mathcal{S}-\eta\,\nabla_{\mathcal{S}}\mathcal{L}$ \Comment{Adam}
\EndFor
\State \Return $\mathcal{S}$ \Comment{decode with $G$ only downstream}
\end{algorithmic}
\end{algorithm}

\paragraph{MAPA hyperparameters.}
Table~\ref{tab:supp_mapa_hparam_global} lists the global MAPA
hyperparameters used in all experiments. Table~\ref{tab:supp_mapa_hparam_data}
lists the dataset-specific anchor and storage settings. The byte-level
ratio $\rho$ is measured from the input and latent tensor shapes and
determines the storage-equivalent prototype count. The spatial
downsampling factor $f$ is shown only for reference.

\begin{table}[t]
\centering
\caption{Per-dataset autoencoder geometry and storage-equivalent prototype
budget. The prototype count per class is $\mathrm{IPC}\cdot\rho$.}
\label{tab:supp_budget}
\scriptsize
\setlength{\tabcolsep}{3.8pt}
\begin{tabular}{lccccc}
\toprule
Dataset & Volume & Latent & $f$ & $\rho$ & Proto./class \\
\midrule
MosMedData & $1{\times}96^3$ & $16{\times}12^3$ & \tnum{8} & \tnum{32} & $\mathrm{IPC}{\times}32$ \\
MedMNIST3D & $1{\times}28^3$ & $16{\times}7^3$ & \tnum{4} & \tnum{4} & $\mathrm{IPC}{\times}4$ \\
MRNet & $1{\times}128^3$ & $64{\times}8^3$ & \tnum{16} & \tnum{64} & $\mathrm{IPC}{\times}64$ \\
\bottomrule
\end{tabular}
\end{table}

\begin{table}[t]
\centering
\caption{Global MAPA hyperparameters.}
\label{tab:supp_mapa_hparam_global}
\setlength{\tabcolsep}{5pt}
\renewcommand{\arraystretch}{1.08}
\begin{tabular}{lc}
\toprule
Hyperparameter & Value \\
\midrule
Prototype optimizer & Adam \\
Prototype learning rate & \tnum{5\ensuremath{\times}10\textsuperscript{-3}} \\
Distillation steps $T$ & \tnum{100} \\
Class/allocation scale $\lambda_{\mathrm{dist}}$ & \tnum{0.2} \\
Anti-collapse weight $\lambda_b$ & \tnum{1.0} \\
Assignment temperature $\tau$ & \tnum{0.05} \\
Curriculum & cosine, \tnum{0} to \tnum{0.7}$T$ \\
Prototype initialization & sampled real latents \\
Random feature extractor $\phi$ & resampled each step \\
\bottomrule
\end{tabular}
\end{table}

\begin{table}[t]
\centering
\caption{Dataset-specific MAPA settings.}
\label{tab:supp_mapa_hparam_data}
\setlength{\tabcolsep}{4pt}
\renewcommand{\arraystretch}{1.08}
\begin{tabular}{lcccc}
\toprule
Dataset & $f$ & $\rho$ & $K$ & $\lambda_m^{\max}$ \\
\midrule
MosMedData & \tnum{8} & \tnum{32} & \tnum{8} & \tnum{0.03} \\
OrganMNIST3D & \tnum{4} & \tnum{4} & \tnum{4} & \tnum{0.10} \\
NoduleMNIST3D & \tnum{4} & \tnum{4} & \tnum{4} & \tnum{0.10} \\
VesselMNIST3D & \tnum{4} & \tnum{4} & \tnum{4} & \tnum{0.10} \\
MRNet & \tnum{16} & \tnum{64} & \tnum{2} & \tnum{0.03} \\
\bottomrule
\end{tabular}
\end{table}

\paragraph{Autoencoder configuration.}
Each dataset uses a lightweight residual 3D convolutional autoencoder
trained on the target training split and then frozen. The encoder and
decoder are reused by all latent baselines and MAPA across all IPC
settings. Table~\ref{tab:supp_ae_config}
summarizes the logged architecture settings and training objective, and
Table~\ref{tab:supp_ae_cost} reports the measured training cost and
reconstruction metrics.

\begin{table*}[t]
\centering
\captionsetup{width=\textwidth}
\caption{Autoencoder configuration from the training logs. Loss terms use
the weights shown in the last column.}
\label{tab:supp_ae_config}
\scriptsize
\setlength{\tabcolsep}{3.5pt}
\renewcommand{\arraystretch}{1.08}
\begin{tabular}{lcccccc}
\toprule
Dataset & Input shape & Latent shape & Channels & Block & Params & Training objective \\
\midrule
MosMedData & $1{\times}96{\times}96{\times}96$ & $16{\times}12{\times}12{\times}12$ & \tnum{16,32,64} & residual & \tnum{0.597M} &
$\mathcal{L}_1+0.05\,\mathrm{MSE}+0.02\,\mathrm{Grad}+0.03\,\mathrm{SSIM}$ \\
OrganMNIST3D & $1{\times}28{\times}28{\times}28$ & $16{\times}7{\times}7{\times}7$ & \tnum{32,64} & residual & \tnum{0.524M} &
$\mathcal{L}_1+0.05\,\mathrm{MSE}+0.02\,\mathrm{Grad}+0.05\,\mathrm{SSIM}$ \\
NoduleMNIST3D & $1{\times}28{\times}28{\times}28$ & $16{\times}7{\times}7{\times}7$ & \tnum{32,64} & residual & \tnum{0.524M} &
$\mathcal{L}_1+0.05\,\mathrm{MSE}+0.02\,\mathrm{Grad}+0.05\,\mathrm{SSIM}$ \\
VesselMNIST3D & $1{\times}28{\times}28{\times}28$ & $16{\times}7{\times}7{\times}7$ & \tnum{16,32} & residual & \tnum{0.132M} &
$\mathcal{L}_1+0.10\,\mathrm{MSE}$ \\
MRNet & $1{\times}128{\times}128{\times}128$ & $64{\times}8{\times}8{\times}8$ & \tnum{16,32,64,128} & residual & \tnum{2.464M} &
$\mathcal{L}_1+0.10\,\mathrm{MSE}+0.01\,\mathrm{Grad}$ \\
\bottomrule
\end{tabular}
\end{table*}

\paragraph{Downstream evaluation configuration.}
Unless otherwise specified, all distilled sets and baselines are evaluated
with the same ConvNet3D classifier protocol. The classifier width is $32$,
the evaluation seed is $101$, and models are trained for $100$ epochs with
SGD, learning rate $0.01$, momentum $0.9$, batch size $8$, and no weight
decay. The selected checkpoint is the one with the best validation
balanced accuracy, and test balanced accuracy is the primary metric. The
cross-architecture experiments below keep the same
synthetic sets and replace only the downstream classifier.

\subsection{Cross-Architecture Readout}
\label{subsec:supp_cross_arch}

On two downstream architectures distinct from the default ConvNet3D, namely
ResNet18-3D~\cite{supp:tran2018closer} and
MobileNetV2-3D~\cite{supp:Sandler_2018_CVPR} (MosMedData IPC${=}1$ and $10$,
Table~\ref{tab:cross_arch_mobilenet}), MAPA has the strongest mean
bAcc at every budget. The margin is large at IPC${=}10$ on ResNet18-3D
($57.53$ versus\ $49.29$) and clear on MobileNetV2-3D ($59.17$ versus\
$57.78$), with directional gains at IPC${=}1$. This indicates that the
benefit is not tied to the default readout.

\begin{table}[!t]
\centering
\captionsetup{skip=2pt}
\caption{Cross-architecture MosMedData bAcc. The distilled sets are
unchanged. Bold denotes the best result, and the MAPA row is shaded.}
\label{tab:cross_arch_mobilenet}
\scriptsize
\setlength{\tabcolsep}{2.4pt}
\renewcommand{\arraystretch}{0.88}
\begin{tabular}{lcccc}
\toprule
& \multicolumn{2}{c}{ResNet18-3D~\cite{supp:tran2018closer}} & \multicolumn{2}{c}{MobileNetV2-3D~\cite{supp:Sandler_2018_CVPR}} \\
\cmidrule(lr){2-3}\cmidrule(lr){4-5}
Method & IPC${=}1$ & IPC${=}10$ & IPC${=}1$ & IPC${=}10$ \\
\midrule
Latent-MMD~\cite{supp:JMLR:v13:gretton12a} & \ms{57.83}{3.09} & \ms{49.29}{1.00} & \ms{61.04}{2.15} & \ms{57.44}{3.78} \\
Latent-DC~\cite{supp:zhao2021dc} & \ms{57.52}{4.58} & \ms{49.50}{1.44} & \ms{62.24}{1.45} & \ms{56.54}{5.64} \\
Latent-DM~\cite{supp:zhao2023dm} & \ms{56.91}{3.73} & \ms{48.42}{2.29} & \ms{61.69}{0.95} & \ms{57.78}{4.04} \\
Latent-MTT~\cite{supp:cazenavette2022mtt} & \ms{58.19}{4.02} & \ms{48.69}{2.93} & \ms{61.86}{0.84} & \ms{55.34}{2.39} \\
\maparow \textbf{MAPA} & \best{\ms{58.29}{3.37}} & \best{\ms{57.53}{3.87}} & \best{\ms{62.43}{2.74}} & \best{\ms{59.17}{4.54}} \\
\bottomrule
\end{tabular}
\end{table}

\subsection{Latent Selection versus Optimization}
\label{subsec:supp_latent_selection}

A natural alternative is to select real latents directly under the same
storage-equivalent budget. We compare against Latent-Medoids, which takes
per-class K-means medoids of real latents under the same autoencoder,
prototype count, decoder, and evaluation as MAPA
(Table~\ref{tab:latent_selection}). This strong baseline approaches MAPA on
simpler near-saturated tasks within seed variation (for example, NoduleMNIST3D and
VesselMNIST3D at IPC${=}1$). Nonetheless, MAPA improves in all 11 settings,
with clear margins on harder ones ($+5.04$ on MosMedData IPC${=}10$, $+4.65$
on VesselMNIST3D IPC${=}10$, and $+7.64$ on MRNet IPC${=}10$). Together with
the anchor-source ablation in the main paper,
this shows that the gain requires real intra-class anchors and optimized
allocation, not selection alone.

\begin{table}[t]
\centering
\captionsetup{width=\linewidth}
\caption{Latent selection versus optimization (bAcc, seeds 42 to 46).
Latent-Medoids selects the K-means medoids of real latents per class under
the same storage-equivalent budget, autoencoder, decoder, and evaluation
protocol as MAPA. MAPA improves over this strong selection baseline in all
11 settings, with the largest margins on harder settings, while on simpler
near-saturated tasks the gap is within seed variation. This isolates the
benefit of optimizing the allocation beyond selecting real latents.}
\label{tab:latent_selection}
\scriptsize
\setlength{\tabcolsep}{2.6pt}
\renewcommand{\arraystretch}{1.04}
\resizebox{\linewidth}{!}{%
\begin{tabular}{llccc}
\toprule
Dataset & IPC & Latent-Medoids & MAPA & $\Delta$ \\
\midrule
MosMedData & \tnum{1} & \ms{63.51}{3.69} & \ms{64.92}{3.57} & \tnum{+1.41} \\
MosMedData & \tnum{10} & \ms{68.38}{3.65} & \ms{73.42}{2.45} & \tnum{+5.04} \\
MosMedData & \tnum{20} & \ms{71.09}{2.34} & \ms{73.03}{5.42} & \tnum{+1.94} \\
OrganMNIST3D & \tnum{1} & \ms{81.64}{2.78} & \ms{82.36}{1.05} & \tnum{+0.72} \\
OrganMNIST3D & \tnum{10} & \ms{92.42}{0.90} & \ms{93.37}{0.95} & \tnum{+0.95} \\
NoduleMNIST3D & \tnum{1} & \ms{62.22}{2.32} & \ms{62.53}{1.50} & \tnum{+0.31} \\
NoduleMNIST3D & \tnum{10} & \ms{71.88}{2.01} & \ms{72.05}{2.62} & \tnum{+0.17} \\
VesselMNIST3D & \tnum{1} & \ms{65.96}{4.72} & \ms{66.07}{1.67} & \tnum{+0.11} \\
VesselMNIST3D & \tnum{10} & \ms{70.92}{1.03} & \ms{75.57}{3.99} & \tnum{+4.65} \\
MRNet & \tnum{1} & \ms{55.92}{3.25} & \ms{70.76}{5.12} & \tnum{+14.84} \\
MRNet & \tnum{10} & \ms{72.99}{3.21} & \ms{80.63}{1.35} & \tnum{+7.64} \\
\bottomrule
\end{tabular}
}
\end{table}

\begin{table}[t]
\centering
\caption{Matched-IPC MosMedData comparison (bAcc). Latent methods use the
same per-class count as voxel methods rather than their storage-equivalent
budget. This is a control, so best entries are not highlighted. At matched
IPC 1, MAPA is below several baselines as expected because $K'{=}1$ disables
multi-anchor allocation.}
\label{tab:mosmed_matched_ipc}
\begingroup
\scriptsize
\setlength{\tabcolsep}{4.0pt}
\renewcommand{\arraystretch}{0.90}
\begin{tabular}{@{}lcc@{}}
\toprule
Method & \multicolumn{2}{c}{MosMedData} \\
\cmidrule(lr){2-3}
& Matched IPC=1 & Matched IPC=10 \\
\midrule
3D-DC~\cite{supp:zhao2021dc} & \ms{57.01}{4.86} & \ms{57.49}{4.42} \\
Latent-DC~\cite{supp:zhao2021dc} & \ms{56.33}{5.84} & \ms{56.70}{5.76} \\
\midrule
3D-DM~\cite{supp:zhao2023dm} & \ms{61.04}{6.28} & \ms{58.13}{5.05} \\
Latent-DM~\cite{supp:zhao2023dm} & \ms{57.76}{5.93} & \ms{60.67}{6.17} \\
\midrule
3D-MTT~\cite{supp:cazenavette2022mtt} & \ms{58.47}{5.67} & \ms{58.22}{6.48} \\
Latent-MTT~\cite{supp:cazenavette2022mtt} & \ms{57.00}{5.54} & \ms{58.37}{8.47} \\
\midrule
\textbf{MAPA} & \ms{56.91}{4.69} & \ms{61.59}{4.74} \\
\bottomrule
\end{tabular}
\endgroup
\end{table}

%% file: sec/supp_concentration.tex
\section{Prototype Allocation Visualization}
\label{sec:supp_concentration}

\begin{figure}[!t]
\centering
\includegraphics[width=\linewidth]{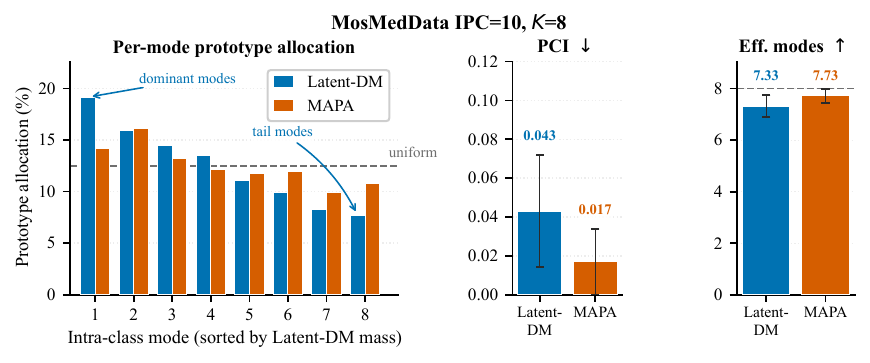}
\caption{Per-anchor prototype allocation on MosMedData (IPC${=}10$, $K{=}8$).
Anchors are sorted by Latent-DM mass. MAPA spreads the same prototype budget
more evenly across anchors, including under-covered tail anchors.}
\label{fig:mode_coverage}
\end{figure}

Figure~\ref{fig:mode_coverage} visualizes the per-anchor prototype allocation
on MosMedData (IPC${=}10$, $K{=}8$). The extended qualitative comparison below
further shows the nearest real slice and decoded prototypes for dominant and
MAPA-covered tail anchors.

\begin{figure*}[!t]
\centering
\includegraphics[width=0.86\textwidth]{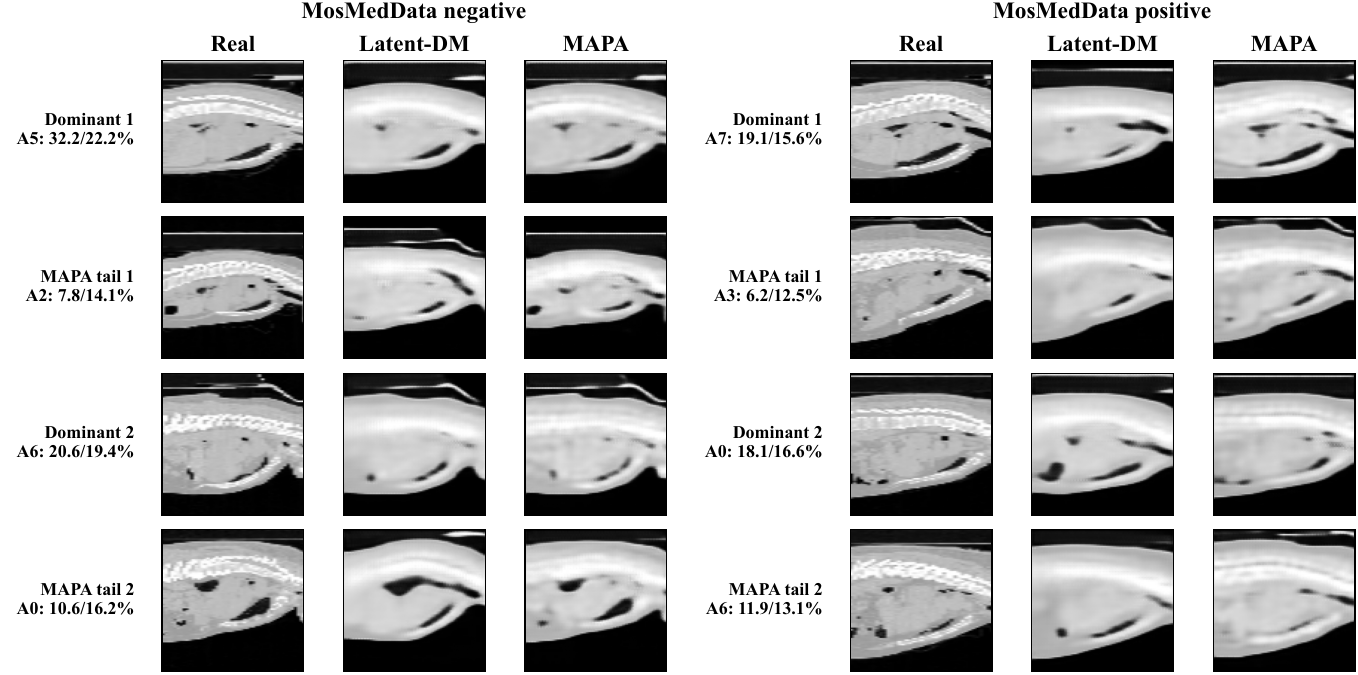}
\captionsetup{width=\textwidth}
\caption{Extended anchor-conditioned qualitative comparison on MosMedData
(IPC${=}10$, $K{=}8$). Each row corresponds to a real-latent anchor. We show
the nearest real slice, the nearest Latent-DM decoded prototype, and the
nearest MAPA decoded prototype to the same anchor. Row labels report the
anchor index and prototype mass as Latent-DM/MAPA percentages. For example,
A5 denotes anchor 5, and 32.2/22.2\% denotes the percentage of prototypes
assigned to that anchor by Latent-DM and MAPA. Dominant rows
correspond to anchors with high Latent-DM mass, while MAPA tail rows
correspond to anchors where MAPA increases allocation relative to Latent-DM.
The visualization qualitatively illustrates that MAPA reallocates prototypes
from dominant anchors toward under-covered intra-class latent regions while
preserving plausible decoded structure. These are representative anchors
selected for visualization and are not a quantitative claim about clinical
subtypes or severity.}
\label{fig:supp_anchor_qualitative}
\end{figure*}

\begin{table}[!t]
\centering
\caption{Prototype concentration across all five datasets at IPC=10
(seeds 42 to 46). Lower PCI and higher normalized EffModes indicate more
balanced allocation. Accuracy is reported in the main tables, and PCI is
used as a coverage diagnostic rather than a direct accuracy predictor.
Absolute concentration levels vary across tasks, so the table supports a
consistent direction of coverage improvement rather than identical PCI levels.}
\label{tab:pci_all_datasets}
\scriptsize
\setlength{\tabcolsep}{4.0pt}
\resizebox{\linewidth}{!}{%
\begin{tabular}{lcccc}
\toprule
Dataset & Latent-DM PCI & MAPA PCI & Latent-DM EffModes & MAPA EffModes \\
\midrule
MosMedData & \tnum{4.30} & \tnum{1.70} & \tnum{91.62} & \tnum{96.62} \\
OrganMNIST3D & \tnum{18.18} & \tnum{13.41} & \tnum{78.68} & \tnum{83.79} \\
NoduleMNIST3D & \tnum{10.28} & \tnum{1.00} & \tnum{86.92} & \tnum{98.64} \\
VesselMNIST3D & \tnum{6.56} & \tnum{3.55} & \tnum{91.50} & \tnum{95.24} \\
MRNet & \tnum{7.86} & \tnum{0.10} & \tnum{94.70} & \tnum{99.93} \\
\bottomrule
\end{tabular}
}
\end{table}

\section{Mode-Count Sensitivity}
\label{sec:supp_k_sensitivity}

\begin{figure*}[!t]
\centering
\includegraphics[width=0.62\textwidth]{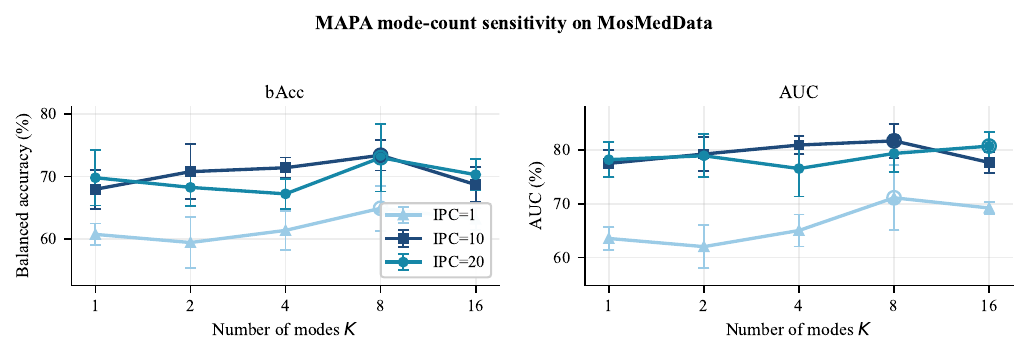}
\captionsetup{width=\textwidth}
\caption{Mode-count sensitivity of MAPA on MosMedData (balanced accuracy
and AUC). $K=8$ reuses the main-table setting for each IPC. Other values
come from the dedicated sensitivity sweep.}
\label{fig:mapa_k_ablation}
\end{figure*}

Figure~\ref{fig:mapa_k_ablation} reports the sensitivity of MAPA to the
anchor count $K\in\{1,2,4,8,16\}$ on MosMedData across IPC $\in\{1,10,20\}$.
The $K{=}8$ points reuse the main-table MAPA setting, while the remaining
$K$ values come from the dedicated sweep. Performance is reasonably stable
across neighboring values, with $K{=}8$ remaining a strong default.

\begin{table*}[!t]
\centering
\captionsetup{width=\textwidth}
\caption{Global $K$ sweep on all datasets (bAcc, seeds 42 to 46). We compare
Latent-DM against MAPA under a single anchor count $K$ shared across all
datasets ($K{=}2,4,8$), versus the per-dataset default $K$ used in the main
results. ``invalid'' denotes $K>N_c$, where anchors outnumber prototypes.
The default $K$ matches or exceeds Latent-DM in all settings and is the
strongest MAPA choice in 10 of 11 settings. The only exception is MRNet
IPC${=}1$, where $K{=}4$ is higher, while the default $K{=}2$ remains above
Latent-DM. When $K$ departs strongly from the intra-class variation of a
dataset, the gain weakens, consistent with $K$ being intended to reflect intra-class
variation rather than a tuned hyperparameter.}
\label{tab:global_k_sweep}
\scriptsize
\setlength{\tabcolsep}{4.2pt}
\renewcommand{\arraystretch}{1.05}
\resizebox{0.92\textwidth}{!}{%
\begin{tabular}{lcccccc}
\toprule
Dataset & IPC & Default $K$ & Latent-DM & MAPA $K{=}2$ & MAPA $K{=}4$ & MAPA $K{=}8$ \\
\midrule
MosMedData & \tnum{1} & \tnum{8} & \ms{60.69}{1.74} & \ms{61.38}{2.97} & \ms{60.60}{4.05} & \ms{64.92}{3.57} \\
MosMedData & \tnum{10} & \tnum{8} & \ms{69.64}{3.09} & \ms{69.04}{4.90} & \ms{72.37}{4.24} & \ms{73.42}{2.45} \\
MosMedData & \tnum{20} & \tnum{8} & \ms{69.75}{5.13} & \ms{70.42}{6.15} & \ms{66.15}{4.73} & \ms{73.03}{5.42} \\
\midrule
OrganMNIST3D & \tnum{1} & \tnum{4} & \ms{78.40}{2.63} & \ms{77.79}{3.55} & \ms{82.36}{1.05} & invalid \\
OrganMNIST3D & \tnum{10} & \tnum{4} & \ms{91.48}{1.71} & \ms{92.73}{1.85} & \ms{93.37}{0.95} & \ms{92.39}{1.72} \\
\midrule
NoduleMNIST3D & \tnum{1} & \tnum{4} & \ms{52.32}{3.78} & \ms{58.15}{7.50} & \ms{62.53}{1.50} & invalid \\
NoduleMNIST3D & \tnum{10} & \tnum{4} & \ms{70.94}{5.79} & \ms{70.65}{4.23} & \ms{72.05}{2.62} & \ms{68.94}{2.63} \\
\midrule
VesselMNIST3D & \tnum{1} & \tnum{4} & \ms{57.49}{4.18} & \ms{61.00}{4.87} & \ms{66.07}{1.67} & invalid \\
VesselMNIST3D & \tnum{10} & \tnum{4} & \ms{71.11}{5.78} & \ms{73.00}{4.95} & \ms{75.57}{3.99} & \ms{71.70}{3.74} \\
\midrule
MRNet & \tnum{1} & \tnum{2} & \ms{69.20}{6.66} & \ms{70.76}{5.12} & \ms{73.98}{2.42} & \ms{71.73}{4.07} \\
MRNet & \tnum{10} & \tnum{2} & \ms{77.94}{2.91} & \ms{80.63}{1.35} & \ms{76.86}{4.43} & \ms{76.34}{6.86} \\
\bottomrule
\end{tabular}
}
\end{table*}

\section{Prototype-Count Scaling}
\label{sec:supp_proto_count_scaling}

Table~\ref{tab:supp_mosmed_proto_count_scaling} varies the prototype count
per class ($16$ versus\ $32$) on MosMedData. MAPA benefits most from the larger
count ($67.82$ to $73.42$ bAcc at IPC${=}10$), while class-level latent
baselines gain little or can degrade. This confirms that the enlarged set of
latent prototypes is most useful when it is explicitly allocated across
real-latent anchors.

\begin{table*}[!t]
\centering
\captionsetup{width=\textwidth}
\caption{Prototype-count scaling on MosMedData (balanced accuracy). The MAPA rows are shaded.}
\label{tab:supp_mosmed_proto_count_scaling}
\scriptsize
\setlength{\tabcolsep}{4.8pt}
\renewcommand{\arraystretch}{1.08}
\begin{tabular}{lccc}
\toprule
Method & Proto./class & \multicolumn{2}{c}{MosMedData} \\
\cmidrule(lr){3-4}
& & IPC1 & IPC10 \\
\midrule
\multirow{2}{*}{Latent-MMD~\cite{supp:JMLR:v13:gretton12a}}
& \tnum{16} & \ms{59.28}{3.03} & \ms{68.17}{2.28} \\
& \tnum{32} & \ms{61.91}{3.70} & \ms{64.83}{6.99} \\
\midrule
\multirow{2}{*}{Latent-DC~\cite{supp:zhao2021dc}}
& \tnum{16} & \ms{60.27}{5.59} & \ms{68.66}{3.54} \\
& \tnum{32} & \ms{60.02}{3.12} & \ms{68.23}{3.50} \\
\midrule
\multirow{2}{*}{Latent-DM~\cite{supp:zhao2023dm}}
& \tnum{16} & \ms{60.51}{4.72} & \ms{66.60}{3.67} \\
& \tnum{32} & \ms{60.69}{1.74} & \ms{69.64}{3.09} \\
\midrule
\multirow{2}{*}{Latent-MTT~\cite{supp:cazenavette2022mtt}}
& \tnum{16} & \ms{60.20}{5.18} & \ms{68.78}{3.87} \\
& \tnum{32} & \ms{62.69}{2.10} & \ms{64.41}{4.60} \\
\midrule
\maparow \textbf{MAPA}
& \tnum{16} & \ms{61.98}{3.61} & \ms{67.82}{2.46} \\
\maparow
& \tnum{32} & \ms{64.92}{3.57} & \ms{73.42}{2.45} \\
\bottomrule
\end{tabular}
\end{table*}

%% file: sec/supp_ae.tex
\section{Autoencoder and Runtime Accounting}
\label{sec:supp_ae}

The autoencoder is trained once per dataset and then frozen. The same
encoder and decoder pair is reused by all latent baselines and MAPA across
all prototype budgets. It is not retrained for each distillation run. This
section separates the one-time autoencoder cost, distillation-stage cost,
decode/export cost, downstream classifier training cost, and qualitative
reconstruction behavior.

\subsection{Autoencoder Training Cost}
\label{subsec:supp_ae_train}

Table~\ref{tab:supp_ae_cost} reports the autoencoder training records for
all five datasets.

\begin{table*}[t]
\centering
\captionsetup{width=\textwidth}
\caption{Autoencoder accounting from training logs. Time per epoch
is computed as total training time divided by the best epoch. Peak memory
is measured during autoencoder training.}
\label{tab:supp_ae_cost}
\setlength{\tabcolsep}{5pt}
\renewcommand{\arraystretch}{1.08}
\begin{tabular}{lcccccc}
\toprule
Dataset & Best epoch & Total time & Time/epoch & Peak memory & L1 & MSE \\
\midrule
MosMedData & \tnum{8} & \tnum{38.92\,s} & \tnum{4.86\,s} & \tnum{0.91\,GB} & \tnum{0.0280} & \tnum{0.00359} \\
OrganMNIST3D & \tnum{30} & \tnum{6.01\,s} & \tnum{0.20\,s} & \tnum{0.57\,GB} & \tnum{0.0855} & \tnum{0.01486} \\
NoduleMNIST3D & \tnum{30} & \tnum{6.55\,s} & \tnum{0.22\,s} & \tnum{0.57\,GB} & \tnum{0.0365} & \tnum{0.00301} \\
VesselMNIST3D & \tnum{30} & \tnum{7.04\,s} & \tnum{0.23\,s} & \tnum{0.17\,GB} & \tnum{0.00347} & \tnum{0.00192} \\
MRNet & \tnum{5} & \tnum{66.85\,s} & \tnum{13.37\,s} & \tnum{0.78\,GB} & \tnum{0.0372} & \tnum{0.00419} \\
\bottomrule
\end{tabular}
\end{table*}

\subsection{Runtime Accounting at IPC=10}
\label{subsec:supp_runtime_accounting}

Table~\ref{tab:supp_accounting} reports the IPC=10 runtime accounting. For
latent methods, the autoencoder cost is added once, while decode/export and
downstream training are reported separately. For voxel-space methods, the
table includes the completed 3D-DM, 3D-DC, and 3D-MTT runs when available,
and records out-of-memory failures as experimental outcomes.

\begin{table*}[t]
\centering
\captionsetup{width=\textwidth}
\caption{IPC=10 accounting with downstream training included. The
autoencoder is trained once per dataset and reused by all latent methods.
Its cost is therefore added once to Latent-DM and MAPA. Distillation,
decode/export, downstream training, and evaluation times are averaged over
seeds 42 to 46 unless noted. The longer downstream training for latent
methods comes from decoding the storage-equivalent latent budget into more
training volumes and is shared by all latent methods. OOM indicates that
all five seeds failed during distillation on the 96\,GB GPU. The MAPA rows
are shaded.}
\label{tab:supp_accounting}
\scriptsize
\setlength{\tabcolsep}{2.2pt}
\renewcommand{\arraystretch}{1.08}
\resizebox{\linewidth}{!}{%
\begin{tabular}{llccccccc}
\toprule
Dataset & Method & AE train & \shortstack{Distill\\total/step} & Decode/export &
\shortstack{Downstream\\train} & Val/test &
\shortstack{Total\\incl. AE} & \shortstack{Peak mem.\\distill/eval} \\
\midrule
MosMedData & 3D-DM & -- & \tnum{268.54\,s / 0.269\,s} & -- & \tnum{41.52\,s} & \tnum{59.63\,s} & \tnum{369.74\,s} & \tnum{19.06\,GB / --} \\
 & 3D-DC & -- & \tnum{263.76\,s / 2.638\,s} & -- & \tnum{34.98\,s} & \tnum{95.15\,s} & \tnum{358.91\,s} & \tnum{46.01\,GB / --} \\
 & 3D-MTT & -- & \tnum{1179.90\,s / 5.900\,s} & -- & \tnum{42.58\,s} & \tnum{102.66\,s} & \tnum{1283.95\,s} & \tnum{73.04\,GB / --} \\
 & Latent-DM & \tnum{38.92\,s} & \tnum{1.01\,s / 0.010\,s} & \tnum{46.35\,s} & \tnum{969.17\,s} & \tnum{59.64\,s} & \tnum{1121.88\,s} & \tnum{1.23\,GB / 8.76\,GB} \\
\maparow & MAPA & \tnum{38.92\,s} & \tnum{1.32\,s / 0.013\,s} & \tnum{46.24\,s} & \tnum{971.65\,s} & \tnum{60.51\,s} & \tnum{1127.67\,s} & \tnum{1.23\,GB / 8.76\,GB} \\
\midrule
OrganMNIST3D & 3D-DM & -- & \tnum{36.42\,s / 0.0364\,s} & -- & \tnum{6.07\,s} & \tnum{3.16\,s} & \tnum{45.69\,s} & \tnum{2.35\,GB / --} \\
 & 3D-DC & -- & \tnum{28.46\,s / 0.2846\,s} & -- & \tnum{5.59\,s} & \tnum{8.56\,s} & \tnum{37.02\,s} & \tnum{5.51\,GB / --} \\
 & 3D-MTT & -- & \tnum{140.29\,s / 0.7014\,s} & -- & \tnum{5.66\,s} & \tnum{8.58\,s} & \tnum{148.93\,s} & \tnum{9.75\,GB / --} \\
 & Latent-DM & \tnum{5.98\,s} & \tnum{1.57\,s / 0.0157\,s} & \tnum{1.59\,s} & \tnum{21.90\,s} & \tnum{3.15\,s} & \tnum{35.88\,s} & \tnum{0.49\,GB / 0.82\,GB} \\
\maparow & MAPA & \tnum{5.98\,s} & \tnum{2.27\,s / 0.0227\,s} & \tnum{1.59\,s} & \tnum{22.05\,s} & \tnum{3.17\,s} & \tnum{37.61\,s} & \tnum{0.49\,GB / 0.82\,GB} \\
\midrule
NoduleMNIST3D & 3D-DM & -- & \tnum{8.91\,s / 0.0089\,s} & -- & \tnum{1.50\,s} & \tnum{2.75\,s} & \tnum{13.19\,s} & \tnum{1.01\,GB / --} \\
 & 3D-DC & -- & \tnum{6.45\,s / 0.0645\,s} & -- & \tnum{1.25\,s} & \tnum{3.79\,s} & \tnum{10.23\,s} & \tnum{1.55\,GB / --} \\
 & 3D-MTT & -- & \tnum{21.41\,s / 0.1070\,s} & -- & \tnum{1.45\,s} & \tnum{3.96\,s} & \tnum{25.43\,s} & \tnum{1.88\,GB / --} \\
 & Latent-DM & \tnum{6.56\,s} & \tnum{0.61\,s / 0.0061\,s} & \tnum{0.29\,s} & \tnum{4.26\,s} & \tnum{2.73\,s} & \tnum{16.10\,s} & \tnum{0.35\,GB / 0.77\,GB} \\
\maparow & MAPA & \tnum{6.56\,s} & \tnum{0.78\,s / 0.0078\,s} & \tnum{0.28\,s} & \tnum{4.27\,s} & \tnum{2.76\,s} & \tnum{17.17\,s} & \tnum{0.35\,GB / 0.77\,GB} \\
\midrule
VesselMNIST3D & 3D-DM & -- & \tnum{9.00\,s / 0.0090\,s} & -- & \tnum{1.50\,s} & \tnum{3.06\,s} & \tnum{13.57\,s} & \tnum{1.03\,GB / --} \\
 & 3D-DC & -- & \tnum{6.45\,s / 0.0645\,s} & -- & \tnum{1.25\,s} & \tnum{4.06\,s} & \tnum{10.50\,s} & \tnum{1.56\,GB / --} \\
 & 3D-MTT & -- & \tnum{21.44\,s / 0.1072\,s} & -- & \tnum{1.45\,s} & \tnum{4.31\,s} & \tnum{25.81\,s} & \tnum{1.89\,GB / --} \\
 & Latent-DM & \tnum{6.91\,s} & \tnum{0.65\,s / 0.0065\,s} & \tnum{0.12\,s} & \tnum{4.29\,s} & \tnum{3.09\,s} & \tnum{16.61\,s} & \tnum{0.35\,GB / 0.78\,GB} \\
\maparow & MAPA & \tnum{6.91\,s} & \tnum{0.77\,s / 0.0077\,s} & \tnum{0.12\,s} & \tnum{4.20\,s} & \tnum{3.05\,s} & \tnum{17.31\,s} & \tnum{0.35\,GB / 0.78\,GB} \\
\midrule
MRNet & 3D-DM & -- & \tnum{650.12\,s / 0.650\,s} & -- & \tnum{97.04\,s} & \tnum{101.15\,s} & \tnum{848.42\,s} & \tnum{47.63\,GB / --} \\
 & 3D-DC & -- & \multicolumn{6}{c}{OOM during distillation (5/5 seeds)} \\
 & 3D-MTT & -- & \multicolumn{6}{c}{OOM during distillation (5/5 seeds)} \\
 & Latent-DM & \tnum{66.85\,s} & \tnum{3.14\,s / 0.0314\,s} & \tnum{217.00\,s} & \tnum{4572.29\,s} & \tnum{100.37\,s} & \tnum{4979.08\,s} & \tnum{3.26\,GB / 21.32\,GB} \\
\maparow & MAPA & \tnum{66.85\,s} & \tnum{4.43\,s / 0.0443\,s} & \tnum{216.87\,s} & \tnum{4561.23\,s} & \tnum{99.25\,s} & \tnum{4969.51\,s} & \tnum{3.26\,GB / 21.32\,GB} \\
\bottomrule
\end{tabular}
}
\end{table*}

%% file: sec/supp_significance.tex
\section{Aggregate and Seed-Paired Checks}
\label{sec:supp_significance}

Table~\ref{tab:supp_aggregate_rank} summarizes the storage-equivalent latent
comparison across all task-budget settings in the main and extension
tables, covering MosMedData IPC $\{1,10,20\}$ and the four extension datasets at IPC
$\{1,10\}$, for $11$ settings in total. Mean bAcc is the unweighted average
over settings, and average rank is computed within the latent methods in
each setting, with lower rank better.

\begin{center}
\centering
\captionof{table}{Aggregate bAcc and average rank across all storage-equivalent
latent-method settings ($11$ settings). Bold denotes the best result, and
the MAPA row is shaded.}
\label{tab:supp_aggregate_rank}
\scriptsize
\setlength{\tabcolsep}{5.0pt}
\begin{tabular}{lccc}
\toprule
Method & Mean bAcc & Avg. rank $\downarrow$ & Wins \\
\midrule
Latent-MMD & \tnum{69.89} & \tnum{3.00} & \tnum{0/11} \\
Latent-DC & \tnum{68.91} & \tnum{4.09} & \tnum{0/11} \\
Latent-MTT & \tnum{68.81} & \tnum{3.82} & \tnum{0/11} \\
Latent-DM & \tnum{69.91} & \tnum{3.09} & \tnum{0/11} \\
\maparow \textbf{MAPA} & \best{\tnum{74.06}} & \best{\tnum{1.00}} & \best{\tnum{11/11}} \\
\bottomrule
\end{tabular}
\end{center}

To assess whether the main gains persist under seed-paired comparison, we
compare MAPA against the strongest latent baseline in each selected
setting. For each seed $s\in\{42,\ldots,46\}$, we compute
$\Delta_s=\mathrm{bAcc}^{\mathrm{MAPA}}_s-\mathrm{bAcc}^{\mathrm{base}}_s$.
Table~\ref{tab:supp_significance} reports the mean paired difference, a
paired bootstrap 95\% confidence interval obtained by resampling the five
seed pairs, and an exact two-sided sign-flip permutation $p$ value. With
only five seed pairs, the two-sided sign-flip permutation test has a
minimum attainable two-sided $p$ value of $2\cdot 2^{-5}=0.0625$, so we report $p$
values as descriptive and rely primarily on the confidence intervals to
characterize the gains, following the broader statistical recommendation to
emphasize effect estimates and their precision rather than significance
thresholds, which is particularly appropriate in small-sample regimes. These comparisons use five seed pairs. Runtime is reported in
Supplementary~Table~\ref{tab:supp_accounting}. We further note that $\Delta$ is
positive in all six evaluated settings, indicating a consistent direction of
improvement.

\begin{center}
\centering
\captionof{table}{Seed-paired bAcc differences between MAPA and the strongest
latent baseline in selected settings. Positive $\Delta$ favors MAPA.}
\label{tab:supp_significance}
\scriptsize
\setlength{\tabcolsep}{3.0pt}
\renewcommand{\arraystretch}{1.08}
\resizebox{\columnwidth}{!}{%
\begin{tabular}{llcccc}
\toprule
Setting & Baseline & MAPA & Baseline & $\Delta$ & 95\% CI / $p$ \\
\midrule
MosMedData IPC1 & Latent-MTT & \tnum{64.92} & \tnum{62.69} & \tnum{+2.23} & \tnum{[-1.24, +4.65] / 0.3125} \\
MosMedData IPC10 & Latent-DM & \tnum{73.42} & \tnum{69.64} & \tnum{+3.78} & \tnum{[+0.89, +6.16] / 0.1250} \\
MosMedData IPC20 & Latent-MTT & \tnum{73.03} & \tnum{71.78} & \tnum{+1.25} & \tnum{[-3.96, +4.54] / 0.6875} \\
OrganMNIST3D IPC10 & Latent-DM & \tnum{93.37} & \tnum{91.48} & \tnum{+1.89} & \tnum{[+1.21, +2.56] / 0.0625} \\
NoduleMNIST3D IPC10 & Latent-MMD & \tnum{72.05} & \tnum{71.56} & \tnum{+0.49} & \tnum{[-3.19, +4.70] / 0.6875} \\
VesselMNIST3D IPC10 & Latent-MMD & \tnum{75.57} & \tnum{74.13} & \tnum{+1.45} & \tnum{[-3.82, +5.79] / 0.7500} \\
\bottomrule
\end{tabular}
}
\end{center}